\documentclass{article}
\usepackage{lmodern}
\usepackage[english]{babel}

\usepackage[letterpaper,top=1cm,bottom=1.2cm,left=2cm,right=2cm,marginparwidth=1.75cm]{geometry}

\usepackage{amsmath}
\usepackage{graphicx}
\usepackage[colorlinks=false, allcolors=blue]{hyperref}
\usepackage{natbib}
\usepackage{amsthm,amssymb,bm}
\usepackage{nicefrac}
\usepackage{cancel}
\usepackage{relsize}
\usepackage{xspace}
\usepackage[most]{tcolorbox}
\usepackage{float}
\usepackage{lmodern}

\allowdisplaybreaks

\newcommand{\acro}[1]{\textsc{#1}\xspace}

\newcommand{\utab}{Tab.\xspace}
\newcommand{\usec}{Sec.\xspace}
\newcommand{\eqn}{Eq.\xspace}
\newcommand{\eqns}{Eqs.\xspace}
\newcommand{\ie}{i.e.\xspace}
\newcommand{\eg}{e.g.\xspace}
\newcommand{\wrt}{w.r.t\xspace}
\newcommand{\etc}{e.t.c\xspace}
\newcommand{\KLD}{\acro{\small KLD}}

\newcommand{\cbox}[1]{\begin{tcolorbox}[colback=gray!10!white, colframe=gray!50!black, boxrule=0.5pt]#1\end{tcolorbox}}
\newcommand{\mysum}[2]{\sum_{#1}^{#2}}

\newcommand{\phivec}{\bm{\phi}}
\newcommand{\phimat}{\bm{\Phi}}

\newcommand{\real}{\mathbb{R}}
\newcommand{\realN}{\mathbb{R}^N}
\newcommand{\realNrow}{\mathbb{R}^{1\times N}}
\newcommand{\realM}{\mathbb{R}^M}
\newcommand{\realNM}{\mathbb{R}^{N\times M}}
\newcommand{\realNN}{\mathbb{R}^{N\times N}}
\newcommand{\realNsquare}{\mathbb{R}^{N^2}}
\newcommand{\realNNsquare}{\mathbb{R}^{N\times N^2}}
\newcommand{\realNsquareN}{\mathbb{R}^{N^2\times N}}
\newcommand{\realNsquareNsquare}{\mathbb{R}^{N^2\times N^2}}
\newcommand{\realNsquarerow}{\mathbb{R}^{1\times N^2}}
\newcommand{\realNDuplicationrow}{\mathbb{R}^{1\times \nicefrac{1}{2}(N^2 + N)}}
\newcommand{\realNDup}{\mathbb{R}^{\nicefrac{1}{2}(N^2 + N)}}

\newcommand{\realNNDup}{\mathbb{R}^{N\times \nicefrac{1}{2}(N^2 + N)}}
\newcommand{\realNDupN}{\mathbb{R}^{\nicefrac{1}{2}(N^2 + N) \times N}}
\newcommand{\realNDupNDup}{\mathbb{R}^{\nicefrac{1}{2}(N^2 + N) \times \nicefrac{1}{2}(N^2 + N)}}

\newcommand{\realNHessian}{\mathbb{R}^{(2N + 2N^2) \times (2N + 2N^2)}}
\newcommand{\realNHessianDuplication}{\mathbb{R}^{(3N + N^2) \times (3N + N^2)}}

\newtheorem{proposition}{Proposition}

\newcommand{\pare}[1]{\left(#1\right)}
\newcommand{\pareT}[1]{\left(#1\right)^T}
\newcommand{\bra}[1]{\left[#1\right]}
\newcommand{\braT}[1]{\left[#1\right]^T}

\newcommand{\ddsymbol}{\text{d}}
\newcommand{\vecsymbol}{\text{vec}}
\newcommand{\vechsymbol}{\text{vech}}
\newcommand{\tracesymbol}{\text{tr}}

\newcommand{\Amat}{\mathbf{A}}
\newcommand{\avec}{\mathbf{a}}
\newcommand{\Bmat}{\mathbf{B}}
\newcommand{\bvec}{\mathbf{b}}

\newcommand{\Cmat}{\mathbf{C}}
\newcommand{\cvec}{\mathbf{c}}

\newcommand{\Dmat}{\mathbf{D}}
\newcommand{\dvec}{\mathbf{d}}

\newcommand{\Xmat}{\mathbf{X}}
\newcommand{\Xmatinv}{\mathbf{X}^{-1}}
\newcommand{\xvec}{\mathbf{x}}

\newcommand{\Ymat}{\mathbf{Y}}
\newcommand{\yvec}{\mathbf{y}}

\newcommand{\zvec}{\mathbf{z}}

\newcommand{\Imat}{\mathbf{I}}

\newcommand{\Dn}{\Dmat_n}
\newcommand{\DnT}{\Dmat_n^T}

\newcommand{\mvec}{\mathbf{m}}
\newcommand{\wvec}{\mathbf{w}}

\newcommand{\Smat}{\mathbf{S}}
\newcommand{\Smatt}{\mathbf{S}^T}
\newcommand{\Smatinv}{\mathbf{S}^{-1}}
\newcommand{\Smatinvt}{\pareT{\mathbf{S}^{-1}}}
\newcommand{\Vmat}{\mathbf{V}}
\newcommand{\Vmatt}{\mathbf{V}^T}
\newcommand{\Vmatinv}{\mathbf{V}^{-1}}
\newcommand{\Vmatinvt}{\pareT{\mathbf{V}^{-1}}}

\newcommand{\uvec}{\mathbf{u}}

\newcommand{\zeromat}{\mathbf{0}}

\newcommand{\dd}[1]{\ddsymbol #1}
\newcommand{\ddB}[1]{\ddsymbol \left[#1\right]}
\newcommand{\Bdd}[1]{\left[\ddsymbol #1\right]}
\newcommand{\Tdd}[1]{\left[\ddsymbol #1\right]^T}

\newcommand{\ddT}[1]{\ddsymbol #1^T}

\newcommand{\vvec}[1]{\vecsymbol\,#1}
\newcommand{\vvecB}[1]{\vecsymbol \left[#1\right]}
\newcommand{\vvecP}[1]{\vecsymbol \left(#1\right)}

\newcommand{\vvech}[1]{\vechsymbol\,#1}

\newcommand{\vvechP}[1]{\vechsymbol \left(#1\right)}

\newcommand{\vvecBT}[1]{\left[\vvecB{#1} \right]^T}
\newcommand{\vvecT}[1]{\left[ \vvec{#1}\right]^T}
\newcommand{\vvechT}[1]{\left[ \vvech{#1}\right]^T}

\newcommand{\tr}[1]{\tracesymbol #1}
\newcommand{\trB}[1]{\tracesymbol \left[#1\right]}

\newcommand{\Jacobian}[1]{J_{#1}}
\newcommand{\Hessian}[1]{H_{#1}}

\newcommand{\circularTR}{\trB{\Amat\Bmat\Xmat} = \trB{\Xmat\Amat\Bmat}}
\newcommand{\scalarTR}{ a \in  \real \Rightarrow a = \trB{a}}
\newcommand{\linearTR}{\trB{\Amat + \Bmat} = \trB{\Amat}+\trB{\Bmat}}
\newcommand{\quadraticScalar}{ \xvec^T\Bmat\xvec \in \real}
\newcommand{\scalarTranspose}{ a \in  \real \Rightarrow a = a^T}

\newcommand{\prodTranspose}{\pare{\Amat\Xmat\Bmat}^T = \Bmat^T\Xmat^T\Amat^T}
\newcommand{\sumTranspose}{\pareT{\Amat+\Bmat} = \Amat^T + \Bmat^T}

\newcommand{\kronTranspose}{\pare{\Amat \otimes \Bmat}^T = \Amat^T \otimes \Bmat^T}

\newcommand{\kronDistribut}{\pare{\Amat+\Bmat}\otimes \Cmat = \Amat \otimes \Cmat + \Bmat\otimes\Cmat}

\newcommand{\kronScalar}{\pare{\alpha\Amat \otimes \beta\Bmat} = \alpha\beta\pare{\Amat\otimes\Bmat}}

\newcommand{\symetricTranspose}{\Amat = \Amat^T}
\newcommand{\transposeInverse}{\pare{\Amat^{-1}}^T = \pare{\Amat^T}^{-1}}
\newcommand{\matprodIdentity}{\Amat=\Amat\Imat}

\newcommand{\blockTranspose}{
\begin{bmatrix}
\Amat &  \Bmat\\
\Cmat & \Dmat
\end{bmatrix}^T = 
\begin{bmatrix}
\Amat^T &  \Cmat^T\\
\Bmat^T & \Dmat^T
\end{bmatrix}
}

\title{\textbf{The Jacobian and Hessian of the Kullback-Leibler Divergence between Multivariate Gaussian Distributions \\} Technical Report }
\author{{\huge Juan Maroñas}\\\vspace{0.75cm} {\normalsize \textit{CUNEF Universidad}} \vspace{-1cm}}

\begin{document}
\maketitle

\begin{abstract}
     This document shows how to obtain the Jacobian and Hessian matrices of the Kullback-Leibler divergence between two multivariate Gaussian distributions, using the first and second-order differentials. The presented derivations are based on the theory presented by \cite{magnus99}. I've also got great inspiration from some of the derivations in \cite{minka}.

    Since I pretend to be at most didactic, the document is split into a summary of results and detailed derivations on each of the elements involved, with specific references to the tricks used in the derivations, and to many of the underlying concepts.
\end{abstract}

 All the derivations presented in this work are obviously inspired by some of the tricks or final expressions\footnote{In the sense of looking forward to ways of deriving expressions that yield a solution that has a particular form, so that checking properties, such as symmetry, is straightforward.} provided in both \cite{magnus99,minka}, but are not a mere copy of them, \ie they have been explicitly derived. Thus, the reader might find new tricks, solutions to new examples, or more details about some tricks used in the derivations from (mainly) \cite{minka}. The reader might also find similar expressions to those in the examples of \cite{minka} since many different expressions involving Gaussian distributions (\KLD, log-likelihood \etc) have terms in common (up to different signs, or different multiplying constants). These, in conjunction with obtaining the presented results through different paths, have been a sanity check towards ensuring correct derivations. Also,  for consistency, while \cite{minka} uses the canonical form of the first differential of a matrix-valued scalar function $\phi:\realNN \to \real$ in the form of:
 \begin{align*}
     \dd y = \tr[\Amat\dd\Xmat],
 \end{align*}
where the Jacobian is identified by $\Amat$; I've preferred using the standard practice of transforming a matrix into vectors through the $\vvec{}$ operator and expressing the first-order differential through:
 \begin{align*}
     \dd y = \vvecT{\Amat}\dd\vvec\Xmat
 \end{align*}
with the Jacobian being $\vvecT{\Amat}$. 
The only difference is whether the Jacobian is expressed as a matrix or a vector. This distinction stems from the fact that \cite{minka} starts by identifying which types of functions yield Jacobians that can naturally be represented as matrices. In contrast, \cite{magnus99} argues that, in the context of matrix differentiation, it is rather more convenient to express matrices through vectors and then use standard vector calculus. I have found this viewpoint more convenient for my interests.

Finally, while I have consciously checked all the steps involved in the derivations, it might be the case that some of them might be missing or do not apply at that specific point, especially with the differential of a linear projection $\ddB{\Amat \Xmat} = \Amat \dd{\Xmat}$. Also, while I express second-order derivatives as:
\begin{align*}
    \frac{\dd \phi(\xvec)}{\dd \xvec^T \dd\xvec}
\end{align*}
the second order differential is expressed using:
\begin{align*}
    \dd^2 \phi(\xvec) = \Tdd{\xvec} \Bmat \dd\xvec
\end{align*}
Here the reader shall note that since $\dd{\xvec^T} = \Tdd{\xvec}$ then both ways are equivalent. I have used the same notation as \cite{magnus99} who expresses differentials and partial derivatives using this notation.
\tableofcontents
\section{Summary of results}

Whoever is familiar enough with the theory under the presented derivations can skip the whole paper and read just this section.

Consider two $N$-dimensional multivariate Gaussian distributions:
\begin{align*}
    p(\xvec) &= \mathcal{N}\left(\xvec\mid\wvec,\Vmat\right)\\
    q(\xvec) &= \mathcal{N}\left(\xvec\mid\mvec,\Smat\right)
\end{align*}
then:
\begin{align*}
    \KLD[q(\xvec)\mid\mid p(\xvec)] &= \frac{1}{2}\bra{\log\mid\Vmat\mid - \log \mid\Smat\mid -N +\trB{\Vmatinv\Smat}+(\mvec-\wvec)^T\Vmatinv(\mvec-\wvec)}
\end{align*}
with $\mvec,\wvec \in \realN$ and $\Vmat,\Smat \in \mathbb{S}^N_{++} \subset \realNN$, \ie the subspace of positive definitive matrices of dimension $N\times N$. We know that covariance matrices are symmetric, which implies $\Smat=\Smatt,\Vmat=\Vmatt$. Denote as $\Dn \in \real^{N^2 \times \nicefrac{1}{2}(N^2 + N)}$ the duplication matrix, which is used to select the unique elements from $\Smat,\Vmat$, \ie the elements from the lower triangular submatrices, see \citep{magnus99,minka} or \usec \ref{vec_operator}.
\subsection{Jacobian}
\begin{align*}
    \Jacobian{\KLD} = \begin{bmatrix}
        \frac{\dd \KLD}{\dd \mvec^T}  & \frac{\dd \KLD}{\dd \wvec^T}& \frac{\dd \KLD}{\dd \vvecT{\Smat}}& \frac{\dd \KLD}{\dd \vvecT{\Vmat}}  
    \end{bmatrix} \in \real^{1 \times 2N+2N^2}
\end{align*}
\begin{align*}
     \frac{\dd \KLD}{\dd \mvec^T} &= (\mvec-\wvec)^T\Vmatinv  && \in \realNrow\\
     \frac{\dd \KLD}{\dd \wvec^T} &= (\wvec-\mvec)^T\Vmatinv  && \in \realNrow\\
     \frac{\dd \KLD}{\dd \vvecT{\Smat}} &= \frac{1}{2}\vvecBT{\Vmatinv-\Smatinv}  && \in \realNsquarerow\\
     \frac{\dd \KLD}{\dd \vvecT{\Vmat}} &= \frac{1}{2}\vvecBT{\Vmatinv-\Vmatinv\pare{\Smat+(\mvec-\wvec)(\mvec-\wvec)^T}\Vmatinv} && \in \realNsquarerow
\end{align*}
Since optimization usually runs on unique elements, it is very common to parameterize covariance matrices through, for example, their Cholesky factorization. However, since this work is mainly about theoretical derivations for further analysis, the reader should be able to obtain specific Jacobians depending on the parameterizations by applying one step further in the Chain Rule. 

In any case, one can use the matrix theory underlying these derivations to express the Jacobian through its unique elements using the Duplication matrix. These are given by:
\begin{align*}
    \Jacobian{\KLD} = \begin{bmatrix}
        \frac{\dd \KLD}{\dd \mvec^T}  & \frac{\dd \KLD}{\dd \wvec^T}& \frac{\dd \KLD}{\dd \vvechT{\Smat}}& \frac{\dd \KLD}{\dd \vvechT{\Vmat}}  
    \end{bmatrix} \in \real^{1 \times 3N+N^2}
\end{align*}
with the new terms being:
\begin{align*}
     \frac{\dd \KLD}{\dd \vvechT{\Smat}} &= \frac{1}{2}\vvecBT{\Vmatinv-\Smatinv}\Dn  && \in \realNDuplicationrow\\
     \frac{\dd \KLD}{\dd \vvechT{\Vmat}} &= \frac{1}{2}\vvecBT{\Vmatinv-\Vmatinv\pare{\Smat+(\mvec-\wvec)(\mvec-\wvec)^T}\Vmatinv}\Dn && \in \realNDuplicationrow
\end{align*}
\subsection{Hessian}

\begin{align*}
    \Hessian{\KLD} = \begin{bmatrix}
\frac{\dd \KLD}{\dd \mvec^2} & \frac{\dd \KLD}{\dd \mvec^T\,\dd\wvec} & \frac{\dd \KLD}{\dd \mvec^T\,\dd\vvec{\Smat}} & \frac{\dd \KLD}{\dd \mvec^T\,\dd\vvec{\Vmat}} \\\\
\frac{\dd \KLD}{\dd \wvec^T\,\dd \mvec} & \frac{\dd \KLD}{\dd \wvec^2} & \frac{\dd \KLD}{\dd \wvec^T\,\dd\vvec{\Smat}} & \frac{\dd \KLD}{\dd \wvec^T\,\dd\vvec{\Vmat}} \\\\
\frac{\dd \KLD}{\dd\vvecT{\Smat}\,\dd \mvec} & \frac{\dd \KLD}{\dd\vvecT{\Smat}\,\dd\wvec} & \frac{\dd \KLD}{\dd\vvec{\Smat}^2} & \frac{\dd \KLD}{\dd\vvecT{\Smat}\,\dd\vvec{\Vmat}} \\\\
\frac{\dd \KLD}{\dd\vvecT{\Vmat}\,\dd \mvec} & \frac{\dd \KLD}{\dd\vvecT{\Vmat}\,\dd\wvec} & \frac{\dd \KLD}{\dd\vvecT{\Vmat}\,\dd\vvec{\Smat}} & \frac{\dd \KLD}{\dd\vvec{\Vmat}^2}
\end{bmatrix} \in \realNHessian
\end{align*}
\begin{align*}
\frac{\dd \KLD}{\dd \mvec^T\,\dd\mvec} &= \Vmatinv \in \realNN \\
\frac{\dd \KLD}{\dd \mvec^T\,\dd\wvec} &= -\Vmatinv  \in \realNN \\
\frac{\dd \KLD}{\dd\mvec^T\,\dd\vvec{\Smat}} &= \zeromat \in \realNNsquare \\
\frac{\dd \KLD}{\dd\mvec^T \dd \vvec\Vmat} &= -\bra{(\mvec-\wvec)^T\Vmatinv}\otimes \Vmatinv \in \realNNsquare \\\\
\frac{\dd \KLD}{\dd \wvec^T\,\dd\mvec} &= -\Vmatinv \in \realNN \\
\frac{\dd \KLD}{\dd \wvec^T\,\dd\wvec} &= \Vmatinv \in \realNN \\
\frac{\dd \KLD}{\dd\wvec^T\,\dd\vvec{\Smat}} &= \zeromat \in \realNNsquare \\
\frac{\dd \KLD}{\dd\wvec^T \dd \vvec\Vmat} &= \bra{(\mvec-\wvec)^T\Vmatinv} \otimes \Vmatinv \in \realNNsquare \\\\
\frac{\dd \KLD}{\dd \vvecT{\Smat}\,\dd\mvec} &= \zeromat \in \realNsquareN \\
\frac{\dd \KLD}{\dd \vvecT{\Smat}\,\dd\wvec} &= \zeromat \in \realNsquareN \\
\frac{\dd \KLD}{\dd \vvecT{\Smat}\,\dd\vvec\Smat} &= \frac{1}{2}\bra{\Smatinv\otimes \Smatinv} \in \realNsquareNsquare \\
\frac{\dd \KLD}{\dd\vvecT{\Smat},\dd\vvec{\Vmat}} &= -\frac{1}{2}\bra{\Vmatinv\otimes \Vmatinv}  \in \realNsquareNsquare \\\\
 \frac{\dd \KLD}{\dd \vvecT{\Vmat}\dd\mvec} &= -\bra{\Vmatinv(\mvec-\wvec)}\otimes\Vmatinv \in \realNsquareN \\
\frac{\dd \KLD}{\dd \vvecT\Vmat\dd\wvec} &= \bra{\Vmatinv(\mvec-\wvec)}\otimes\Vmatinv \in \realNsquareN \\
\frac{\dd \KLD}{\dd\vvecT{\Vmat},\dd\vvec{\Smat}} &= -\frac{1}{2}\bra{\Vmatinv\otimes \Vmatinv } \in \realNsquareNsquare \\
\frac{\dd\KLD}{\dd \vvecT{\Vmat},\dd\vvec\Vmat} &= -\frac{1}{2}\Vmatinv\otimes\bra{\Vmatinv-2\Vmatinv\pare{\Smat+(\mvec-\wvec)(\mvec-\wvec)^T}\Vmatinv} \in \realNsquareNsquare
\end{align*}
As with the Jacobian, we can consider only the Hessian over the unique elements, giving:
\begin{align*}
    \Hessian{\KLD} = \begin{bmatrix}
\frac{\dd \KLD}{\dd \mvec^2} & \frac{\dd \KLD}{\dd \mvec^T\,\dd\wvec} & \frac{\dd \KLD}{\dd \mvec^T\,\dd\vvech{\Smat}} & \frac{\dd \KLD}{\dd \mvec^T\,\dd\vvech{\Vmat}} \\\\
\frac{\dd \KLD}{\dd \wvec^T\,\dd \mvec} & \frac{\dd \KLD}{\dd \wvec^2} & \frac{\dd \KLD}{\dd \wvec^T\,\dd\vvech{\Smat}} & \frac{\dd \KLD}{\dd \wvec^T\,\dd\vvech{\Vmat}} \\\\
\frac{\dd \KLD}{\dd\vvechT{\Smat}\,\dd \mvec} & \frac{\dd \KLD}{\dd\vvechT{\Smat}\,\dd\wvec} & \frac{\dd \KLD}{\dd\vvech{\Smat}^2} & \frac{\dd \KLD}{\dd\vvechT{\Smat}\,\dd\vvech{\Vmat}} \\\\
\frac{\dd \KLD}{\dd\vvechT{\Vmat}\,\dd \mvec} & \frac{\dd \KLD}{\dd\vvechT{\Vmat}\,\dd\wvec} & \frac{\dd \KLD}{\dd\vvechT{\Vmat}\,\dd\vvech{\Smat}} & \frac{\dd \KLD}{\dd\vvech{\Vmat}^2}
\end{bmatrix} \in \realNHessianDuplication,
\end{align*}
with new elements:
\begin{align*}
    \frac{\dd \KLD}{\dd \mvec^T\,\dd\vvech{\Smat}} &= \zeromat\Dn && \in \realNNDup\\
    \frac{\dd \KLD}{\dd \mvec^T\,\dd\vvech{\Vmat}} &= -\bra{(\mvec-\wvec)^T\Vmatinv}\otimes \Vmatinv \Dn && \in \realNNDup\\\\
    \frac{\dd \KLD}{\dd \wvec^T\,\dd\vvech{\Smat}} &= \zeromat \Dn && \in \realNNDup\\
    \frac{\dd \KLD}{\dd \wvec^T\,\dd\vvech{\Vmat}} &= \bra{(\mvec-\wvec)^T\Vmatinv}\otimes \Vmatinv \Dn && \in \realNNDup\\\\\
    \frac{\dd \KLD}{\dd\vvechT{\Smat}\,\dd \mvec} &= \DnT\zeromat && \in \realNDupN\\
    \frac{\dd \KLD}{\dd\vvechT{\Smat}\,\dd\wvec} &= \DnT\zeromat && \in \realNDupN\\
    \frac{\dd \KLD}{\dd\vvech{\Smat}^2} &= \frac{1}{2}\DnT\bra{\Smatinv\otimes\Smatinv}\Dn && \in \realNDupNDup\\
    \frac{\dd \KLD}{\dd\vvechT{\Smat}\,\dd\vvech{\Vmat}} &= -\frac{1}{2}\DnT\bra{\Vmatinv\otimes\Vmatinv}\Dn && \in \realNDupNDup \\\\
    \frac{\dd \KLD}{\dd\vvechT{\Vmat}\,\dd \mvec} &= -\DnT \bra{\Vmatinv(\mvec-\wvec)}\otimes\Vmatinv && \in \realNDupN\\
    \frac{\dd \KLD}{\dd\vvechT{\Vmat}\,\dd\wvec} &= \DnT \bra{\Vmatinv(\mvec-\wvec)}\otimes\Vmatinv && \in \realNDupN\\
    \frac{\dd \KLD}{\dd\vvechT{\Vmat}\,\dd\vvech{\Smat}} &= -\frac{1}{2}\DnT \bra{\Vmatinv\otimes\Vmatinv}\Dn  && \in \realNDupNDup\\
    \frac{\dd \KLD}{\dd\vvech{\Vmat}^2} &=  -\frac{1}{2}\DnT\bra{ \Vmatinv\otimes\bra{\Vmatinv-2\Vmatinv\pare{\Smat+(\mvec-\wvec)(\mvec-\wvec)^T}\Vmatinv}}\Dn  && \in \realNDupNDup
\end{align*}
\section{Preliminaries}
Some introductions to concepts used in this document.

\subsection{Vec operator}
\label{vec_operator}
The $\vecsymbol$ operator receives as input a matrix $\Amat$ and returns a vector containing the columns of the matrix in order, one after the other.
\begin{align*}
\vvecP{
\begin{bmatrix}
a_{11} & a_{12} & a_{13} \\
a_{21} & a_{22} & a_{23} \\
a_{31} & a_{32} & a_{33}
\end{bmatrix} 
}
=
\begin{bmatrix}
a_{11} \\
a_{21} \\
a_{31} \\
a_{12} \\
a_{22} \\
a_{32} \\
a_{13} \\
a_{23} \\
a_{33}
\end{bmatrix} \in \realNsquare
\end{align*}
If $\Amat$ is symmetric, then we similarly define the $\vechsymbol$ operator, which returns only the unique elements of the matrix, \ie, the diagonal and the lower triangular.
\begin{align*}
\vvechP{
\begin{bmatrix}
a_{11} & a_{12} & a_{13} \\
a_{12} & a_{22} & a_{23} \\
a_{13} & a_{23} & a_{33}
\end{bmatrix}
}
=
\begin{bmatrix}
a_{11} \\
a_{12} \\
a_{13} \\
a_{22} \\
a_{23} \\
a_{33}
\end{bmatrix} \in \realNDup
\end{align*}
The relation between both operators is given by:
\begin{align*}
    \Dn \vvech{\Amat} = \vvec{\Amat}
\end{align*}
where $\Dn \in \real^{N^2 \times \nicefrac{1}{2}(N^2 + N)}$ is the Duplication matrix \citep{magnus99,minka}.
\subsection{Jacobian Matrix}

The Jacobian matrix is the matrix of partial derivatives. For a vector-valued function $\phivec : \realN \to \realM$ we  have:
\begin{align}
\Jacobian{\phivec} = \frac{\dd\phivec(\xvec)}{\dd\xvec^T} = 
\begin{bmatrix}
\frac{\partial \phi_1}{\partial x_1} & \frac{\partial \phi_1}{\partial x_2} & \cdots & \frac{\partial \phi_1}{\partial x_N} \\
\frac{\partial \phi_2}{\partial x_1} & \frac{\partial \phi_2}{\partial x_2} & \cdots & \frac{\partial \phi_2}{\partial x_N} \\
\vdots & \vdots & \ddots & \vdots \\
\frac{\partial \phi_M}{\partial x_1} & \frac{\partial f\phi_M}{\partial x_2} & \cdots & \frac{\partial \phi_M}{\partial x_N}
\end{bmatrix}\label{equ:vector_valued_jacobian}
\end{align}
For a scalar function $\phi: \realN \to \real$, the Jacobian is the transpose of the gradient vector.
\begin{align*}
\Jacobian{\phi} = \frac{\dd\phi(\xvec)}{\dd\xvec^T} = 
\begin{bmatrix}
  \frac{\partial \phi}{\partial x_1} & \frac{\partial \phi}{\partial x_2} & \cdots & \frac{\partial \phi}{\partial x_N}\end{bmatrix}  =  \braT{\nabla \phi(\xvec)}
\end{align*}
For a vector-valued scalar argument function $\phivec: \real \to \realM$, the Jacobian is a column vector. 
\begin{align*}
\Jacobian{\phivec} = \frac{\dd\phivec(x)}{\dd x} = 
\begin{bmatrix}
\frac{\partial \phi_1}{\partial x} \\
\frac{\partial \phi_2}{\partial x} \\
\vdots  \\
\frac{\partial \phi_M}{\partial x} 
\end{bmatrix}
\end{align*}
Whenever the domain or the range of the function is a matrix, \eg $\phimat: \realN \to \realNM $, $\phi: \realNM \to \real$, $\phimat: \realNM \to \realNM$, $\phivec: \realNM \to \realN$ (or any possible combination), we will invoke the $\vecsymbol$ or $\vechsymbol$ operators on the matrix and apply standard vector calculus. This is one of the main ideas exposed in \cite{magnus99}: \emph{Since matrix calculus is not so well defined as vector calculus, let's convert matrices to vectors and operate as usual}. 
For example, the Jacobian of a scalar-valued matrix function is a $N \times M$ row vector.
\begin{align*}
\Jacobian{\phi} = \frac{\dd\phi(\Xmat)}{\dd\vvecT{\Xmat}} = 
\begin{bmatrix}
\frac{\partial \phi}{\partial x_1} & \frac{\partial \phi}{\partial x_2} & \cdots & \frac{\partial \phi}{\partial x_{N\times M}}
\end{bmatrix}
\end{align*}
For a matrix valued function $\phimat : \realNM \to \realNN$ we have a $N^2 \times NM$ matrix:
\begin{align*}
\Jacobian{\phimat} = \frac{\dd\vvec\phimat(\Xmat)}{\dd\vvecT{\Xmat}} = 
\begin{bmatrix}
\frac{\partial \phi_1}{\partial x_1} & \frac{\partial \phi_1}{\partial x_2} & \cdots & \frac{\partial \phi_1}{\partial x_{N\times M}} \\
\frac{\partial \phi_2}{\partial x_1} & \frac{\partial \phi_2}{\partial x_2} & \cdots & \frac{\partial \phi_2}{\partial x_{N\times M}} \\
\vdots & \vdots & \ddots & \vdots \\
\frac{\partial \phi_{N\times N}}{\partial x_1} & \frac{\partial \phi_{N\times N}}{\partial x_2} & \cdots & \frac{\partial \phi_{N\times N}}{\partial x_{N\times M}}
\end{bmatrix}
\end{align*}
\subsection{Hessian Matrix}

For the Hessian, we will only consider scalar-valued functions $\phi: \real \to \real$, $\phi: \realN \to \real$ and $\phi: \realNM \to \real$, since in only these cases the Hessian can be written with either a scalar or a matrix, and covers our case of interest. In any case, \cite{magnus99} discusses that constructing the full Hessian for vector or matrix-valued functions is not of practical interest, see section $6.3$ in \cite{magnus99}. In such cases, one computes the Hessian for each output dimension.

The Hessian is the matrix of second-order derivatives. For example for $\phi: \realN \to \real$ we have:
\begin{align*}
\Hessian{\phi} = \frac{\dd^2 \phi(\xvec)}{\dd\xvec^T\,\dd\xvec} =
\begin{bmatrix}
\frac{\partial^2 \phi}{\partial x_1^2} & \frac{\partial^2 \phi}{\partial x_1 \partial x_2} & \cdots & \frac{\partial^2 \phi}{\partial x_1 \partial x_N} \\
\frac{\partial^2 \phi}{\partial x_2 \partial x_1} & \frac{\partial^2 \phi}{\partial x_2^2} & \cdots & \frac{\partial^2 \phi}{\partial x_2 \partial x_N} \\
\vdots & \vdots & \ddots & \vdots \\
\frac{\partial^2 \phi}{\partial x_N \partial x_1} & \frac{\partial^2 \phi}{\partial x_N \partial x_2} & \cdots & \frac{\partial^2 \phi}{\partial x_N^2}
\end{bmatrix}
\end{align*}
The Hessian can be seen as the Jacobian of the transposed Jacobian (the gradient), transposed:
\begin{align*}
    \Hessian{\phi} = \left[\frac{\dd}{\dd\xvec^T}\Jacobian{\phi}^T\right]^T = \left[\frac{\dd}{\dd\xvec^T} \nabla \phi(\xvec)\right]^T
\end{align*}
To see so note that for $\phi: \realN \to \real$ we have:
\begin{align*}
\Jacobian{\phi}^T = \left[\frac{\dd\phi(\xvec)}{\dd \xvec^T}\right]^T = 
\begin{bmatrix}
\frac{\partial \phi}{\partial x_1} \\
\frac{\partial \phi}{\partial x_2} \\
\vdots \\
\frac{\partial \phi}{\partial x_{N}}
\end{bmatrix}
\end{align*}
Thus the transposed Jacobian can be seen as a vector-valued function $\phivec: \realN \to \realN$, and so computing the Jacobian again on such a function yields a matrix as shown in \eqn \ref{equ:vector_valued_jacobian}, where now instead of $\phi_1$ we have $\nicefrac{\partial \phi}{\partial x_1}$ as the coordinates of the output of the function.  In particular, note that:
\begin{align*}
\Jacobian{\nabla \phi(\xvec)} = \frac{\dd}{\dd\xvec^T} \nabla \phi(\xvec) = 
\begin{bmatrix}
\frac{\partial \phi}{\partial x_1^2} & \frac{\partial \phi}{\partial x_2 \partial x_1} & \cdots & \frac{\partial \phi}{\partial x_N \partial x_1} \\
\frac{\partial \phi}{\partial x_1\partial x_2} & \frac{\partial \phi}{\partial x_2^2} & \cdots & \frac{\partial \phi}{\partial x_N \partial x_2} \\
\vdots & \vdots & \ddots & \vdots \\
\frac{\partial \phi}{\partial x_1 \partial x_N} & \frac{\partial \phi}{\partial x_2 \partial x_N} & \cdots & \frac{\partial \phi}{\partial x_N^2}
\end{bmatrix} = \Hessian{\phi}^T
\end{align*}
This connection between the Hessian and the Jacobian is important to highlight since, as I'll show, some parts of the Hessian can be easily obtained by the first differential of the Jacobian, rather than the second differential of the function itself.

Two special cases of the Hessian will appear later in the derivations. First note that when the domain is a subset of $\realNM$ then:
\begin{align*}
\Hessian{\phi} = \frac{\dd^2 \phi(\Xmat)}{\dd \vvecT{\Xmat}\,\dd\vvec{\Xmat}} =
\begin{bmatrix}
\frac{\partial^2 \phi}{\partial x_1^2} & \frac{\partial^2 \phi}{\partial x_1 \partial x_2} & \cdots & \frac{\partial^2 \phi}{\partial x_1 \partial x_{N\times M}} \\
\frac{\partial^2 \phi}{\partial x_2 \partial x_1} & \frac{\partial^2 \phi}{\partial x_2^2} & \cdots & \frac{\partial^2 \phi}{\partial x_2 \partial x_{N\times M}} \\
\vdots & \vdots & \ddots & \vdots \\
\frac{\partial^2 \phi}{\partial x_{N\times M} \partial x_1} & \frac{\partial^2 \phi}{\partial x_{N\times M} \partial x_2} & \cdots & \frac{\partial^2 \phi}{\partial x_{N\times M}^2}
\end{bmatrix}
\end{align*}
On the other side, we might be partitioning our vector space $\mathbb{R}^{N^2 + N}$ into two objects of interest, let's say a vector $\xvec \in \realN$ (the mean) and a matrix $\Xmat \in \realNN$ (the covariance). In such cases, we will see that obtaining the Hessian using differentials is easily done by obtaining the second-order differentials from different submatrices of the full Hessian. Some of these submatrices will condense the partial derivatives \wrt $\xvec$ of the gradient of $\Xmat$. More precisely, we will have:
\begin{align*}
\Hessian{\phi} = \frac{\dd^2 \phi(\Xmat,\xvec)}{\dd \xvec^T\,\dd\vvec{\Xmat}} =
\begin{bmatrix}
\frac{\partial^2 \phi}{\partial x_1^2} & \frac{\partial^2 \phi}{\partial x_1 \partial x_2} & \cdots & \frac{\partial^2 \phi}{\partial x_1 \partial x_{N\times N}} \\
\frac{\partial^2 \phi}{\partial x_2 \partial x_1} & \frac{\partial^2 \phi}{\partial x_2^2} & \cdots & \frac{\partial^2 \phi}{\partial x_2 \partial x_{N\times N}} \\
\vdots & \vdots & \ddots & \vdots \\
\frac{\partial^2 \phi}{\partial x_{N} \partial x_1} & \frac{\partial^2 \phi}{\partial x_{N} \partial x_2} & \cdots & \frac{\partial^2 \phi}{\partial x_{N}\partial x_{N\times N}}
\end{bmatrix} \in \realNNsquare
\end{align*}
This is clearly a submatrix from a full Hessian because $\Hessian{\phi} \neq \Hessian{\phi}^T$ since the matrix is non-square. We know that under some assumed conditions the Hessian is a symmetric matrix (see theorem $6.4$ in \cite{magnus99}), and we will always assume these conditions are satisfied.
\subsection{The first-order differential}
Obtaining the Hessian and the Jacobian is thus concerned with obtaining partial derivatives of different types. However, rather than using the usual rules to take derivatives we have been taught in high school, we will be doing so by using the concept of \emph{differential}. As pointed out by \cite{magnus99}: \emph{"Throughout this chapter, and indeed, throughout this book, we shall emphasize the fundamental idea of a differential rather than that of a derivative as this has large practical and theoretical advantages."}, and so there is no reason to not do so. The reader is pointed to section $5.3$ in \cite{magnus99} for a more precise and formal definition of the first differential. I will rather point out directly to its definition and some intuitive descriptions and examples of its usefulness. 

Consider $\phi : \real \to \real$. Consider two points $x,u \in\real$. The standard way of defining the derivative:
\begin{align*}
    \lim_{u\to 0} \frac{\phi(x+u) - \phi(x)}{u}  = \frac{\dd \phi(x)}{\dd x}
\end{align*}
is equivalent to:
\begin{align}
    \phi(x+u) - \phi(x) = a(x)\cdot u + r(x;u); \,\, \lim_{u\to 0} \frac{r(u)}{u} = 0 \label{equ:derivative_first_order_diferential}
\end{align}
from which we identify the first-order differential as $\dd \phi(x) =  a(x) \cdot u$ and $a(x) = \nicefrac{\dd \phi(x)}{\dd x}$\footnote{Although the differential and the derivative are not the same concept I will abuse notation and use $\dd$ to denote both of them. Most of the time, I'll be referring to the differential, and only when a fraction is used, the reader should understand the derivative.} by the first identification theorem, which I discuss below. As noted by \cite{magnus99} we do not require $u$ to be small to define this differential, although this concept is usually used in the limit $u \to 0$.

As pointed out by \cite{minka,magnus99}, we can see the first differential as the best linear approximation to a function at a point. In fact, is the linear approximation of $\phi(x+u) - \phi(x)$ since it engloves all the terms that are linear in $u$. In an analogous way, for vector-valued functions $\phivec: \realN \to \realM$ we have:
\begin{align}
    \phivec(\xvec+\uvec) - \phivec(\xvec) = \Amat(\xvec)\uvec + r(\xvec;\uvec); \,\, \lim_{\uvec\to 0} \frac{r(\uvec)}{||\uvec||} = 0 \label{equ:derivative_first_order_diferential_vectorvalued}
\end{align}
with the first differential being $\Amat(\xvec)\uvec$.

There are two important results we shall briefly expose. The first one is the uniqueness of the first-order differential (theorem $5.3$ in \cite{magnus99}). The second is the first identification theorem. The uniqueness of the first-order differential implies there is no other expression for a differential that satisfies \eqns \ref{equ:derivative_first_order_diferential}, \ref{equ:derivative_first_order_diferential_vectorvalued}. The first identification theorem identifies the derivative or Jacobian from the first-order differential. In particular:
\begin{align*}
    \Amat(\xvec) &= \Jacobian{\phivec}\\
    a(x) &= \frac{\dd \phi(x)}{\dd x}
\end{align*}
\paragraph{Example 1}
Let's put some examples of this definition. Consider:
\begin{align*}
    \phi(x) &= x^2 \\
    \phi(x+u) & = (x+u)^2 = x^2 + 2xu + u^2\\
    \phi(x+u) - \phi(x) &= 2xu + u^2
\end{align*}
from which we identify $2x$ as being the termed multiplied by $u$ and $r(u)=u^2$. It holds that: 
\begin{align*}
\lim_{u \to 0} \frac{r(u)}{u} = \lim_{u \to 0}  u = 0   
\end{align*}
We can see how the first differential identifies the derivative. Taking the derivative of $x^2$ is rather simple and gives $2x$.
\paragraph{Example 2}
Consider now:
\begin{align*}
    \phi(x) &= (x^2 - x)^2 &&= x^4 - 2x^3 + x^2\\
    \phi(x+u) & = ((x+u)^2 - (x+u))^2 &&= x^4 + 4ux^3 - 2x^3 + 6u^2x^2 - 6ux^2 + x^2 +4u^3x - 6u^2x + 2ux+ u^4 - 2u^3 + u^2 \\
    \phi(x+u) - \phi(x) & &&= 4ux^3 + 6u^2x^2 - 6ux^2+4u^3x-6u^2x+2ux+u^4-2u^3+u^2
\end{align*}
from which we identify terms:
\begin{align*}
    \dd \phi(x) &= (4x^3 -6x^2+ 2x)u\\
    r(u) &= 6u^2x^2+4u^3x-6u^2x +u^4-2u^3+u^2
\end{align*}
where:
\begin{align*}
   &\frac{\dd \phi(x)}{\dd x} = 4x^3 -6x^2+ 2x\\
   &\lim_{u \to 0} \frac{6u^2x^2+4u^3x-6u^2x +u^4-2u^3+u^2}{u} = 0
\end{align*}
We can again compute the derivative by standard high school methods, yielding the same result. 
\paragraph{Example 3:} \label{par:example2}
Similar results hold for vector-valued functions. Consider $\xvec = (x_1,x_2)$  and $\uvec = (u_1,u_2)$, such that $\phi(\xvec) = x_1x_2$. Then:
\begin{align*}
    \phi(\xvec) &= x_1x_2\\
    \phi(\xvec+\uvec) &= (x_1+u_1)(x_2+u_2) = x_1x_2 + x_1u_2 + u_1x_2 + u_1u_2\\
    \phi(\xvec+\uvec) - \phi(\xvec) &= x_1u_2 + u_1x_2 + u_1u_2 
\end{align*}
from which we identify terms:
\begin{align*}
    \dd \phi(\xvec) &= x_2u_1 + x_1u_2 = [x_2,x_1]\cdot \uvec\\
    r(\uvec) &= u_1u_2
\end{align*}
where:
\begin{align*}
    \frac{\dd \phi(\xvec)}{\xvec^T} &= [x_2,x_1]\\
    \lim_{\uvec \to 0} \frac{u_1u_2}{\mid\mid \uvec \mid\mid} &= 0
\end{align*}
being $[x_2,x_1]$, the Jacobian of the transformation. This is, again, a standard high school derivative.

The reader would be now concerned by: \emph{why would one go through this method instead of computing derivatives as we usually do?}. Well, it turns out that for many more complex functions (such as those involving matrices or vectors) it is easier to identify the derivative from the differential. The idea is to compute the differential of the expression and write it down in its canonical form; from which we identify the derivative, by the first identification theorem.

Before putting an example, we need to derive common differentials for many expressions, which can be done in several ways.
\subsection{Differentials rules for some common functions}
When taking $u \to 0$, it is common to use the notation $u = \dd x$, and to note that $\dd \phivec(\xvec) = \phivec(\xvec + \dd\xvec) -\phivec(\xvec)  $. So the expressions of the differentials take the form:
\begin{align*}
    \dd \phi(x) = \frac{\dd \phi(x)}{\dd x} \dd x\\
    \dd \phivec(\xvec) = \Jacobian{\phivec} \dd \xvec
\end{align*}
As noted by \cite{minka}, differential rules can be obtained from the definition by writing down  $\phivec(\xvec+\dd \xvec)- \phivec(\xvec)$ and inspecting the linear part, as we have done previously. 

\paragraph{Example 1:}

\begin{align*}
    &\phi(\xvec,\yvec) = \xvec\yvec\\
    &\phi(\xvec + \dd \xvec,\yvec + \dd \yvec) = (\xvec + \dd \xvec) (\yvec + \dd \yvec) =
    \xvec\yvec +\xvec\dd\yvec + \Bdd{\xvec}\yvec + \dd\xvec\dd\yvec\\
    &\phi(\xvec + \dd \xvec,\yvec + \dd \yvec)-\phi(\xvec,\yvec) = \xvec\dd\yvec + \yvec\dd\xvec + \dd\xvec\dd\yvec
\end{align*}
from which the linear part is identified: $\xvec\dd\yvec + \yvec\dd\xvec$. So:
\begin{align*}
    \dd\bra{\xvec\yvec} = \xvec\dd\yvec + \yvec\dd\xvec.
\end{align*}
\paragraph{Example 2:}
\begin{align*}
    &\phi(\xvec,\yvec) = \xvec+\yvec\\
    &\phi(\xvec + \dd \xvec,\yvec + \dd \yvec) = \xvec + \dd \xvec + \yvec + \dd \yvec\\
    &\phi(\xvec + \dd \xvec,\yvec + \dd \yvec)-\phi(\xvec,\yvec) = \dd\xvec + \dd\yvec
\end{align*}
From which:
\begin{align*}
    \dd\bra{\xvec+\yvec} = \dd\xvec + \dd\yvec
\end{align*}
\paragraph{Example 3:}
\begin{align*}
    &\phi(\xvec) = \Amat\xvec\\
    &\phi(\xvec + \dd \xvec) = \Amat(\xvec + \dd \xvec) = \Amat\xvec + \Amat\dd\xvec\\
    &\phi(\xvec + \dd \xvec)-\phi(\xvec) = \Amat\dd\xvec
\end{align*}
From which:
\begin{align*}
    \dd\bra{\Amat\xvec} = \Amat\dd\xvec
\end{align*}
\paragraph{Example 4:}
\begin{align*}
    &\phi(\xvec) = \Amat\\
    &\phi(\xvec + \dd \xvec) = \Amat\\
    &\phi(\xvec + \dd \xvec)-\phi(\xvec) = 0
\end{align*}
From which:
\begin{align*}
    \dd\bra{\Amat} = 0
\end{align*}
We shall note that while in the previous section, we have derived Jacobians/derivatives from $\phi(\xvec + \dd \xvec)-\phi(\xvec)$, in this section we have derived the differential rule for a function. Thus, from this differential rule, we can get the Jacobian by inspection. In other words, the procedure will be to first operate on the differential, place it in the canonical form, and inspect the Jacobian from it. We'll see this in the following subsection.

The last ingredient before this step is how we can use the chain rule, once basic differentials have been derived, to obtain more complex differentials. For differentials, the chain rule is quite similar (see Theorem $5.8$ in \cite{magnus99}). If we have $z = f(y)$ and $y = g(x)$. Then we have the differentials:
\begin{align*}
    \dd \zvec &= \Amat(\yvec) \dd \yvec\\
    \dd \yvec &= \Bmat(\xvec) \dd \xvec\\
    \dd \zvec &= \Amat(\yvec)\Bmat(\xvec) \dd \xvec
\end{align*}
So it is basically the product of the differentials. The reader can also inspect section $2$ in chapter $8$ in \cite{magnus99} to see how many other differentials are obtained, in a procedure different from the one carried out in this section. 

With all these in mind, a potential list of differential rules is the following:
\begin{align*}
    \dd \Bdd{ \Xmat } &= 0 \\
    \dd \Amat &= 0 \\
    \ddB{\Amat\Xmat} &= \Amat\dd\Xmat \\
    \ddB{\Xmat + \Ymat} &= \dd\Xmat + \dd\Ymat \\
    \ddB{\tr \Xmat} &= \trB{\dd\Xmat} \\
    \ddB{\Xmat\Ymat} &= \Xmat\dd\Ymat + \Bdd{\Xmat}\Ymat \\
    \ddB{\Xmat^{-1}} &= -\Xmat^{-1} \dd \Xmat \Xmat^{-1} \\
    \dd \left|\Xmat\right|  &= \left|\Xmat\right|\trB{\Xmat^{-1}\dd\Xmat} \\
    \dd \log \Xmat &= \frac{1}{\Xmat}\dd\Xmat \\
    \ddB{\Xmat^T} &= \braT{\dd\Xmat} \\
    \ddB{\vvech{\Xmat}} &= \vvech{\dd \Xmat}\\ 
    \ddB{\vvec{\Xmat}} &= \vvec{\dd \Xmat} 
\end{align*}
which will be widely used across the derivations. For operators $*$ that rearrange elements (transpose, vec, vech \etc), it holds $\dd \Xmat^* = \Bdd{\Xmat}^*$ \citep{minka}.
\subsection{Obtaining the derivative from the differential}
\label{subsec:derivative_from_dif}
As mentioned before, the basic idea is to obtain the differential of the expression, operate until the expression is provided in its canonical form (the complete list of canonical forms for the first order differential is given in \utab \ref{tab:firstorderdif}), and infer the Jacobian using the first identification theorem. This is rather simpler than using the standard way of computing Jacobians by computing directly the partial derivatives.
\paragraph{Example 1:} To see so, consider a function $\phi : \real^2 \to \real$, where the canonical form is $\dd \phi(\xvec) = \avec(\xvec)^T\dd \xvec$:
\begin{align*}
    \phi(\xvec) &= \xvec^T \Amat \xvec \\
    \xvec &\in \real^2\\
    \Amat &= \begin{pmatrix}
    1 & 2 \\
    1 & 1 \\
    \end{pmatrix}
\end{align*}
The Jacobian of this function is the partial derivatives \wrt the elements of $\xvec = (x_1,x_2)$. More precisely, we have:
\begin{align*}
\phi(\xvec) = \begin{bmatrix}
    x_1 & x_2
\end{bmatrix}\begin{pmatrix}
    1 & 2 \\
    1 & 1 \\
    \end{pmatrix}\begin{bmatrix}
    x_1 \\
    x_2
\end{bmatrix} = x_1^2 + 3x_1x_2 + x_2^2\\
\Jacobian{\phi} = \begin{bmatrix}
\frac{\partial \phi(\xvec)}{\partial x_1},  \frac{\partial \phi(\xvec)}{\partial x_2}
\end{bmatrix} = \begin{bmatrix}
2x_1 + 3x_2 , 3x_1 + 2x_2
\end{bmatrix}
\end{align*}
However, it is rather simpler to identify the Jacobian by computing the first differential and expressing it in its canonical form:
\begin{align*}
    \dd \phi(\xvec) & = \ddB{ \xvec^T \Amat \xvec} && \\
    &=  \ddB{\xvec^T} \Amat \xvec +   \xvec^T \Amat \dd \xvec && \text{using \eqn}  \ref{dprod}, \ref{dAX}\\
    &=  \xvec^T \Amat^T \dd \xvec  +   \xvec^T \Amat \dd\xvec && \text{using \eqn}  \ref{dXT} \text{ and because } \quadraticScalar,\\
    & && \scalarTranspose,\, \prodTranspose\\
    &=  \xvec^T\bra{\Amat^T + \Amat} \dd\xvec \\
    \Jacobian{\phi} &= \xvec^T\bra{\Amat^T + \Amat} = \bra{x_1,x_2}\begin{bmatrix}
        2 & 3 \\
        3 & 2
    \end{bmatrix} = \bra{2x_1 + 3x_2, 3x_1 + 2x_2} && \text{from first identification theorem.}
\end{align*}
where we identify the Jacobian as the term multiplied by $\dd \xvec$.
\paragraph{Example 2:} Another example could involve the chain rule, which I apply explicitly just to illustrate the expression above, but we will see that we do it implicitly most of the time. Consider $\phi:\realNN \to \real$, with canonical form $\dd \phi(\Xmat) = \vvecT{\Amat}\dd \vvec\Xmat$:
\begin{align*}
    \phi(\Xmat) = \log \left|\Xmat\right|, 
\end{align*}
which is given by a composition of two functions:
\begin{align*}
     \nu(z) &= \log z \\
     z &=  \left|\Xmat\right| 
\end{align*}
Obtain differentials for both expressions:
\begin{align*}
    \dd \nu(z) &= \dd \log z && \\
    &= \frac{1}{z}\dd z && \text{using \eqn} \ref{dlog}\\
    \dd z &= \dd \left|\Xmat\right| && \\
    &= \left|\Xmat\right|\trB{\Xmat^{-1} \dd \Xmat} && \text{using \eqn}  \ref{ddet}\\
    &= \left|\Xmat\right|\vvecBT{[\Xmat^{-1}]^T}\dd\vvec{\Xmat} && \text{using \eqns} \ref{tr_matprod_2vec},\ref{dvec}
\end{align*}
Applying the chain rule gives:
\begin{align*}
    \dd \phi(\Xmat) &= \frac{1}{z} \left|\Xmat\right|\vvecBT{[\Xmat^{-1}]^T}\dd\vvec{\Xmat} \\
    &= \frac{1}{\left|\Xmat\right|}\left|\Xmat\right|\vvecBT{[\Xmat^{-1}]^T}\dd\vvec{\Xmat}\\
    &= \vvecBT{[\Xmat^{-1}]^T}\dd\vvec{\Xmat}
\end{align*}
from which the Jacobian is given by: $\vvecBT{[\Xmat^{-1}]^T}$.
\paragraph{Example 3:} Another example could consider matrix value, matrix argument functions  $\Phi: \realNN \to \realNM$, with canonical form $\dd\vvec\Phi(\Xmat) = \Amat \dd \vvec\Xmat$, which will end up clearly clarifying how useful this approach is:
\begin{align*}
    \Phi(\Xmat) = \Xmatinv\Amat\Xmat
\end{align*}
Since the function is matrix-valued, we need to take vec to convert to vectors and then operate. As we have seen, when either the domain or the range is matrix-valued, we convert through vec to a vector, because that is where the rules of calculus are well defined \cite{magnus99}. Thus, to obtain the differential, we have:
\begin{align*}
    \dd \vvecB{\Phi(\Xmat)} &= \ddB{\vvecB{\Xmatinv\Amat\Xmat}} && \\
    &= \vvecB{\dd\Xmatinv\Amat\Xmat + \Xmatinv\Amat\dd\Xmat}  && \text{using \eqns} \ref{dvec},\ref{dprod},\ref{dAX} \\
    &= \vvecB{-\Xmatinv\dd\Xmat\Xmatinv\Amat\Xmat + \Xmatinv\Amat\dd\Xmat\Imat} && \text{using \eqn} \ref{dinv}, \matprodIdentity \\
    &=-\pare{\braT{\Xmatinv\Amat\Xmat} \otimes \Xmatinv} \dd\vvec\Xmat + \pare{\Imat \otimes \bra{\Xmatinv\Amat}} \dd\vvec{\Xmat} && \text{using \eqns } \ref{vecmat2kron},\ref{vecsum},\ref{dvec} \\
    &= \bra{-\pare{\braT{\Xmatinv\Amat\Xmat} \otimes \Xmatinv} + \Imat \otimes \bra{\Xmatinv\Amat}} \dd\vvec{\Xmat} && 
\end{align*}
from which the Jacobian is given by: $-\pare{\braT{\Xmatinv\Amat\Xmat} \otimes \Xmatinv} + \Imat \otimes \bra{\Xmatinv\Amat} $.
\paragraph{Example 4:} A final example considers:
\begin{align*}
    \Phi(\Xmat) = \log \Xmat,
\end{align*}
then:
\begin{align*}
    \dd \vvec \Phi(\Xmat) &= \dd \vvec \log \Xmat && \\
    &= \vvecB{\dd \log \Xmat}  && \text{using \eqn } \ref{dvec} \\
    &= \vvecB{\frac{1}{\Xmat}\dd \Xmat} && \text{using \eqn } \ref{dlog}\\
    &= \pare{\Imat \otimes \frac{1}{\Xmat}} \dd\vvec{\Xmat} && \text{using \eqns } \ref{vecmat2kron},\ref{dvec}, \matprodIdentity 
\end{align*}
This example emphasizes the difference between the differential of the $\log$ function of a matrix and the actual canonical form from which we can identify the Jacobian.

These four examples (specially the third one) should concern the reader with how useful this technique is, especially when expressions become more complex.
\subsection{Canonical forms}

We have seen so far how to compute derivatives for any function using the concept of differential. The idea is to operate the differential of the expression until expressed through its canonical form, from where the Jacobian is identified by the first identification theorem. Whenever the domain or range is matrix-valued, we would need to express the differential using the vec operator. There are $9$ possible canonical forms \cite{magnus99}:
\begin{table}[H]
\centering
\caption{Canonical forms of the first order differential.}
\label{tab:firstorderdif}
\vspace{0.5cm}
\begin{tabular}{|c|c|c|c|}
\hline
\textbf{Mapping}  & \textbf{Canonical form} & \textbf{Jacobian:} & \textbf{Jacobian dimension} \\
% Scalar valued
\hline
$\phi : \real \to \real$  & $\dd\phi(x) = a \dd x$  &  $\Jacobian{\phi} = a$ & $\real$\\
\hline
$\phi : \realN \to \real$  &  $\dd\phi(\xvec) = \avec^T\dd \xvec$ &  $\Jacobian{\phivec} = \avec^T$ & $\real^{1 \times N }$\\
\hline
$\phi : \realNM \to \real$  &  $\dd\phi(\Xmat) = \vvecT{\Amat}\dd \vvec\Xmat$ & $\Jacobian{\phimat} =  \vvecT{\Amat}$   & $\real^{1 \times (N \times M) }$\\
\hline
% Vector valued
$\phivec : \real \to \realN$  &  $\dd\phivec(x) =  \avec\dd x$ & $\Jacobian{\phivec} = \avec$  & $\real^{N \times 1}$\\
\hline
$\phivec : \realN \to \realM$  &  $\dd\phivec(\xvec) =  \Amat\dd \xvec$ & $\Jacobian{\phivec} = \Amat$ & $\real^{M \times N }$\\
\hline
$\phivec : \realNM \to \realN$  &  $\dd\phivec(\Xmat) =  \Amat\dd \vvec\Xmat$ & $\Jacobian{\phivec} = \Amat$ &  $\real^{N \times (N\times M) \ }$\\
\hline
% Matrix valued
$\phimat : \real \to \realNM$  &  $\dd\vvecB{\phimat(x)} = \vvec{\Amat} \dd x$ &  $\Jacobian{\phimat} = \vvec{\Amat}$ & $\real^{(N \times M) \times 1}$\\
\hline
$\phimat : \realN \to \realNM$  &  $\dd\vvecB{\phimat(\xvec)} = \Amat \dd \xvec$ & $\Jacobian{\phimat} = \Amat$ & $\real^{(N\times M) \times N }$\\
\hline
$\phimat : \realNN \to \realNM$  &    $\dd\vvecB{\phimat(\Xmat)} =  \Amat \dd \vvec\Xmat$ & $\Jacobian{\phimat} = \Amat$ & $\real^{(N\times M)\times(N\times N)}$\\
\hline
\end{tabular}
\end{table}
From these canonical forms, we can see the well-known fact from differential geometry about the Jacobian being the matrix that linearly projects the tangent space on the domain to the tangent space in the range. We can see how the Jacobian has the exact dimensions to project a differential in the domain dimension to a differential in the range dimension. 

\paragraph{Example 1:} For example for our previous function $\phi: \realN \to \real$, given by:
\begin{align*}
    \phi(\xvec) &= \xvec^T \Amat \xvec 
\end{align*}
The differential in its canonical form was:
\begin{align*}
    \dd\phi(\xvec) &= \xvec^T\bra{\Amat^T + \Amat} \dd\xvec \\
    \Jacobian{\phi} &= \xvec^T\bra{\Amat^T + \Amat}      
\end{align*}
where we clearly identify a dot product between the Jacobian $\xvec^T\bra{\Amat^T + \Amat} \in \real^{1\times N}$ and the differential $\dd \xvec \in \real^N$, which result in a projection to the differential in the tangent space of the range $\dd \bra{\xvec^T \Amat \xvec}  \in \real$. This fact gives us a rule to derive the total differential of an expression. 

Suppose we have a vector space $\zvec \in \realM$ that we partition into $\xvec \in \realN$ and $\yvec \in \real^{M-N}$, such that $\zvec^T = [\xvec^T,\yvec^T]$. The differential is given by the concatenation of the differential as well $\dd\zvec^T = \dd[\xvec^T,\yvec^T]$. Suppose we have a function $\phi:\realM \to \real$, then the Jacobian of the full transformation will have dimensions $\real^{1\times M}$, such that:
\begin{align*} 
    \dd \phi(\zvec) = \Jacobian{\phi}\dd\zvec = \mysum{i=1}{M}\Jacobian{\phi,i} \dd z_i,
\end{align*}
since the projection is just the dot product between the differential on the tangent space of the domain and the Jacobian. Note that:
\begin{align*}
    \mysum{i=1}{M}\Jacobian{\phi,i} \dd z_i &= \mysum{i=1}{N}\Jacobian{\phi,i}\dd x_i + \mysum{i=1}{M-N}\Jacobian{\phi,N+i} \dd y_i \\
    &= \Jacobian{\phi,\xvec} \dd \xvec + \Jacobian{\phi,\yvec} \dd \yvec
\end{align*}
So we can write the full projection by adequately partitioning the full Jacobian and considering the differentials on each of the different variables. In other words, we can write the full differential of an expression by computing the differentials \wrt some variables and then summing up the products of the corresponding Jacobians and differentials, in case it is an easier way than concatenating and obtaining the full differential at once. In other words, we have: 
\begin{align*}
    \dd\phi(\xvec,\yvec,\zvec) = \Jacobian{\phi,\xvec,\yvec,\zvec} \dd\braT{\xvec^T,\yvec^T,\zvec^T}  =  \Jacobian{\phi,\xvec}\dd\xvec + \Jacobian{\phi,\yvec}\dd\yvec +\Jacobian{\phi,\zvec}\dd\zvec
\end{align*}
This fact is used by \cite{minka} when writing down \eqn $(122)$. This holds for any domain and substructures, as we just need to index the corresponding rows and columns from the Jacobian and express them through the canonical forms outlined above. We will use this fact to obtain the Jacobian of our objective function. In fact, we have seen previously an example where we did this implicitly. More precisely, example 3 in \usec \ref{par:example2}.

\subsection{The second-order differential}
As the reader might imagine, the second-order differential can be used to compute the Hessian by first obtaining the second-order differential, expressing it in the corresponding canonical form, and inferring the Hessian from the expression. Following \cite{magnus99}, we only consider scalar-valued functions, since for other types we just compute Hessians per dimension. I am not particularly interested in such a case since the \KLD is a scalar-valued function. In this section, I will not provide as many examples and particularizations as for the first-order differential since they are not so easy to see in the way we have done previously.

The second order differential is obtained by the differential of the first differential, with the expression given by (for the particular case of vector-valued domain):
\begin{align*}
    \dd \phi(\xvec) &= \Jacobian{\phi}\uvec\\
    \dd^2 \phi(\xvec) &= \ddB{\Jacobian{\phi}\uvec} = \uvec^T\Bmat_\phi\uvec; 
\end{align*}
The second identification theorem tells us that:
\begin{align*}
    \Hessian{\phi} =\begin{cases}
     \frac{1}{2}(\Bmat_\phi + \Bmat_\phi^T) & \text{ if } \Bmat_\phi \neq \Bmat_\phi^T\\     
     \Bmat_\phi & \text{ if } \Bmat_\phi = \Bmat_\phi^T
    \end{cases}
\end{align*}
Note that, for the case $\Bmat_\phi = \Bmat_\phi^T$, then:
\begin{align*}
    \Bmat_\phi= \frac{1}{2}(\Bmat_\phi + \Bmat_\phi^T) 
\end{align*}
which means we could always say:
\begin{align*}
    \Hessian{\phi} = \frac{1}{2}\pare{\Bmat_\phi + \Bmat_\phi^T }
\end{align*}
We also have a uniqueness theorem and chain rule for the Hessian, \cite{magnus99}. I will not apply the chain rule of the Hessian since the way I will be computing second-order differentials does not require us to do so.

In practice, we will have two options to compute the Hessian. The first is to obtain the differential of the first differential and express it in the corresponding canonical form, from which the Hessian is identified. Another option is to note that the Hessian is just the Jacobian of the transposed Jacobian being transposed. From this definition, we can also obtain the Hessian by computing the first-order differential of the transposed Jacobian and then expressing it in the corresponding canonical form for the first-order differential. From there, the Hessian is identified.

The three canonical forms for the Hessian are given by (again substituting $\uvec$ by $\dd\xvec$):
\begin{table}[H]
\centering
\caption{Canonical forms for the second order differential.}
\label{tab:secondorderdif}
\vspace{0.5cm}
\begin{tabular}{|c|c|c|c|}
\hline
\textbf{Mapping} & \textbf{Canonical form} & \textbf{Hessian} & \textbf{Hessian dimension} \\
\hline
$\phi : \real \to \real$  &  $\dd x\, b\, \dd x$ &  $\Hessian{\phi} = b$ & $\real$\\
\hline
$\phi : \realN \to \real$  &  $\Tdd{\xvec} \Bmat \dd \xvec$  &  $\Hessian{\phi} =\begin{cases}
     \nicefrac{1}{2}(\Bmat_\phi + \Bmat_\phi^T) & \text{ if } \Bmat_\phi \neq \Bmat_\phi^T\\     
     \Bmat_\phi & \text{ if } \Bmat_\phi = \Bmat_\phi^T
    \end{cases}$ & $\realNN$\\
\hline
$\phi : \realNM \to \real$  &  $\Tdd{\vvec{\Xmat}} \Bmat \dd \vvec{\Xmat}$  &  $\Hessian{\phi} =\begin{cases}
     \nicefrac{1}{2}(\Bmat_\phi + \Bmat_\phi^T) & \text{ if } \Bmat_\phi \neq \Bmat_\phi^T\\     
     \Bmat_\phi & \text{ if } \Bmat_\phi = \Bmat_\phi^T
    \end{cases}$ & $\real^{(N+M) \times (N+M)}$ \\
\hline
\end{tabular} 
\end{table}
where we will assume that, in general, the Hessian is a symmetric matrix, \ie, theorem $6.4$ in \cite{magnus99} is satisfied.
Similar to the first-order differential, we can compute the full Hessian by computing submatrices. Again consider $\zvec \in \realM$ that we partition into $\xvec \in \realN$ and $\yvec \in \real^{M-N}$, such that $\zvec^T = [\xvec^T,\yvec^T]$. Consider  function $\phi:\realM \to \real$. The Hessian matrix can be partitioned into $4$ submatrices:
\begin{align*}
    \Hessian{\phi} = \begin{bmatrix}
        \frac{\partial\phi(\xvec,\yvec)}{\partial\xvec^2} & \frac{\phi(\xvec,\yvec)}{\partial\xvec^T\partial\yvec}\\
        \frac{\partial\phi(\xvec,\yvec)}{\partial\yvec^T\partial\xvec} & \frac{\phi(\xvec,\yvec)}{\partial\yvec^2} 
    \end{bmatrix}
\end{align*}
Then it turns out we can express this in the second-order differential canonical forms:
\begin{align*}
    \dd_\xvec^2 \phi(\xvec,\yvec) &= \Tdd{\xvec} \Bmat_{\xvec} \dd \xvec\\
    \dd_\xvec \dd_\yvec\phi(\xvec,\yvec) &= \Tdd{\xvec} \Bmat_{\xvec,\yvec} \dd \yvec\\
    \dd_\yvec \dd_\xvec\phi(\xvec,\yvec) &= \Tdd{\yvec} \Bmat_{\yvec,\xvec} \dd \xvec\\
    \dd_\yvec^2 \phi(\xvec,\yvec) &= \Tdd{\yvec} \Bmat_{\yvec} \dd \yvec
\end{align*}
It is clear enough that the submatrices $\Bmat_{\xvec,\yvec},\Bmat_{\yvec,\xvec}$ will not be, in general, valid Hessian matrices. This is easily proved since if $N \neq N-M$, then $\Bmat_{\xvec,\yvec},\Bmat_{\yvec,\xvec}$  will have a different number of rows and columns; and so there is no way to be symmetric nor to make it. So whenever we compute such differentials, we need to keep in mind that they are just submatrices from the full Hessian.

The good point of this is that we can easily compute the Hessian involving arguments that are matrices or vectors. In such cases, and consider for example $\Xmat,\yvec$, then the submatrix corresponding to the partial derivatives can be expressed in canonical form by:
\begin{align*}
     \dd_\yvec \dd_\Xmat\phi(\Xmat,\yvec) &= \Tdd{\yvec} \Bmat_{\yvec,\xvec} \dd \vvec{\Xmat}
\end{align*}
as I have previously noted in the Hessian section. This is also used by \cite{minka} in \eqns $(123-127)$.

Let's see how to compute the Hessian matrix for some examples, through second-order differential. In any case, many of the things outlined in this section are used in the derivations below, and since most of the machinery is similar to that used for the first-order differential, I will not provide as many examples.

\paragraph{Example 1:} Consider: 
\begin{align*}
    \phi(\xvec) &= \xvec^T \Amat \xvec ,
\end{align*}
the first order differential is given by:
\begin{align*}
   \dd\phi(\xvec) &= \xvec^T\bra{\Amat^T + \Amat} \dd\xvec && \text{obtained in example 1 in \usec \ref{subsec:derivative_from_dif}}
\end{align*}
The second order differential is given by:
\begin{align*}
    \dd\bra{\dd\phi(\xvec)} &= \dd\bra{\xvec^T\bra{\Amat^T + \Amat} \dd\xvec}\\
    &= \bra{\dd\xvec^T\bra{\Amat^T + \Amat}\dd\xvec + \xvec^T\bra{\Amat^T + \Amat}\cancelto{0}{\ddB{\dd\xvec}}}, && \text{using \eqns } \ref{dprod},\ref{d2X},\ref{dAX}\\
    &= \Tdd{\xvec}\bra{\Amat^T + \Amat}\dd\xvec && \text{using \eqn} \ref{dXT}
\end{align*}
from where the Hessian is identified:
\begin{align*}
    \Hessian{\phi} = \Amat^T + \Amat
\end{align*}
because $\Amat^T + \Amat$ is symmetric.

\paragraph{Example 2:} Consider now the case in which we have the same function as in the previous example, but the variables being differentiated are both $\xvec$ and $\Amat$. The Hessian is thus made up of $4$ submatrices:

\begin{align*}
    \Hessian{\phi} = \begin{bmatrix}
\frac{\dd \phi(\xvec,\Amat)}{\dd \xvec^2} & \frac{\dd \phi(\xvec,\Amat)}{\dd \xvec^T\,\dd\vvec{\Amat}} \\
\frac{\dd \phi(\xvec,\Amat)}{\dd\vvecT{\Amat}\,\dd \xvec} & \frac{\dd \phi(\xvec,\Amat)}{\dd\vvecT{\Amat}\,\dd\vvec{\Amat}} 
\end{bmatrix}
\end{align*}
where $\frac{\dd \phi(\xvec,\Amat)}{\dd \xvec^T\,\dd\xvec}$ was computed in our previous example. Next, to obtain $\frac{\dd \phi(\xvec,\Amat)}{\dd\vvecT{\Amat}\,\dd\vvec{\Amat}}$, we first compute the first differential \wrt $\Amat$.
\begin{align*}
 \ddB{\xvec^T\Amat \xvec} &= \xvec^T\dd{\Amat} \xvec && \text{using \eqn} \ref{dAX}\\
 &= \trB{\xvec^T\dd{\Amat} \xvec }  && \text{using } \quadraticScalar, \scalarTR\\
 &= \trB{ \xvec \xvec^T\dd{\Amat} }  && \text{using } \circularTR \\
 &= \vvecBT{\braT{\xvec \xvec^T}}\vvec{\dd{\Amat}}  && \text{using \eqn} \ref{tr_matprod_2vec}\\
 &= \vvecBT{\xvec \xvec^T}\dd\vvec{{\Amat}}  && \text{using \eqn } \ref{dvec},\,\symetricTranspose
\end{align*}
The second differential:
\begin{align*}
    \dd\bra{\dd \xvec^T\Amat\xvec} &= \ddB{\vvecBT{\xvec \xvec^T}\dd\vvec{{\Amat}}} && \\
    &= \ddB{\trB{ \xvec \xvec^T\dd{\Amat} }} && \text{reversing derivation from first differential above} \\
    &= \trB{ \xvec \xvec^T\ddB{\ddB{\Amat}}} && \text{using \eqns } \ref{dTR},\ref{dAX}\\
    &= \zeromat && \text{using \eqn } \ref{d2X}
\end{align*}
Next, we need $\frac{\dd \phi(\xvec,\Amat)}{\dd\vvecT{\Amat}\,\dd \xvec}$. Starting now from the first differential over $\xvec$ we have \footnote{Here I got great inspiration from the trick used in \cite{magnus99} Example $10.3$ page $213$. This example would have been much easier if I had assumed that $\Amat$ is symmetric since this is what we find in the quadratic form of the Gaussian distribution. However, this example is particularly interesting for solving differentials involving a matrix and its transpose.}:
\begin{align*}
     \dd_\Amat\bra{\xvec^T\bra{\Amat^T + \Amat} \dd\xvec} &= \xvec^T\bra{\ddB{\Amat^T} + \dd\Amat} \dd\xvec && \text{using \eqns} \ref{dAX},\ref{dsum}\\
     &=\bra{\xvec^T\braT{\dd\Amat} + \xvec^T\dd\Amat} \dd\xvec && \text{using \eqn} \ref{dXT}\\
     &=\braT{\braT{\xvec^T\braT{\dd\Amat} + \xvec^T\dd\Amat}} \dd\xvec && \\
     &=\braT{\dd\Amat\xvec + \braT{\dd\Amat}\xvec} \dd\xvec && \sumTranspose,\prodTranspose\\
     &=\braT{\vvecB{\dd\Amat\xvec} + \vvecB{\braT{\dd\Amat}\xvec}} \dd\xvec && \text{using \eqn} \ref{vec_of_a_vec}\\
     &=\braT{\vvecB{\dd\Amat\xvec} + \vvecB{\xvec^T\dd\Amat}} \dd\xvec && \text{using \eqn} \ref{vec_of_a_vec},  \prodTranspose\\
     &=\braT{\pare{\xvec^T \otimes \Imat}\dd\vvec{\Amat} + \pare{\Imat \otimes \xvec^T}\dd\vvec{\Amat}} \dd\xvec && \text{using \eqns} \ref{vecmat2kron},\ref{dvec}, \matprodIdentity \\
     &=\braT{\dd\vvec{\Amat}}\bra{\pare{\xvec \otimes \Imat} + \pare{\Imat \otimes \xvec}}\dd\xvec && \text{using } \sumTranspose,\\
     & && \prodTranspose,\kronTranspose,\symetricTranspose
\end{align*}
from where:
\begin{align*}
    \frac{\dd \phi(\xvec,\Amat)}{\dd\vvecT{\Amat}\,\dd \xvec} = \pare{\xvec \otimes \Imat} + \pare{\Imat \otimes \xvec}
\end{align*}
Finally, we find $\frac{\dd \phi(\xvec,\Amat)}{\dd \xvec^T\,\dd\vvec{\Amat}}$. Starting from the first differential over $\Amat$, compute the second differential \wrt $\xvec$.
\begin{align*}
    \dd_\xvec\bra{\ddB{\xvec^T\Amat\xvec}} &= \dd_\xvec\bra{\vvecBT{\xvec \xvec^T}\dd\vvec{{\Amat}}}\\
    &= \dd_\xvec\bra{\trB{ \xvec \xvec^T\dd{\Amat} }}  && \text{reversing derivation from first differential above} \\
    &= \trB{ \dd\xvec \xvec^T\dd{\Amat} + \xvec \dd\bra{\xvec^T}\dd{\Amat}} && \text{using \eqns } \ref{dTR},\ref{dprod},\ref{dAX}\\
    &= \trB{ \xvec^T\dd{\Amat}\dd\xvec + \dd\xvec^T\dd{\Amat}\xvec} && \text{using } \linearTR,\circularTR\\
    &= \trB{ \xvec^T\dd{\Amat}\dd\xvec + \xvec^T\braT{\dd{\Amat}}\dd\xvec} && \text{using \eqn }\ref{dXT},\quadraticScalar,\scalarTranspose,\prodTranspose \\
    &= \trB{ \dd\xvec\bra{\xvec^T\dd{\Amat} + \xvec^T\braT{\dd{\Amat}}}} && \text{using } \linearTR,\circularTR
    \\
    &= \vvecT{\dd\xvec} \vvecB{\xvec^T\dd\Amat + \xvec^T\braT{\dd\Amat}} && \text{using \eqns } \ref{tr_matprod_2vec},\ref{vec_of_a_vec}\\
    &= \vvecT{\dd\xvec} \vvecB{\xvec^T\dd\Amat + \dd\Amat\xvec} && \text{using \eqns } \ref{vec_of_a_vec},\ref{vecsum},\prodTranspose\\
    &= \braT{\dd\xvec}\bra{\pare{\Imat \otimes \xvec^T} + \pare{\xvec^T\otimes\Imat}}\dd\vvec{\Amat} && \text{using \eqns } \ref{vec_of_a_vec},\ref{vecsum},\ref{vecmat2kron},\ref{dvec},\matprodIdentity
\end{align*}
from where:
\begin{align*}
    \frac{\dd \phi(\xvec,\Amat)}{\dd \xvec^T\,\dd\vvec{\Amat}} =\pare{\Imat \otimes \xvec^T} + \pare{\xvec^T\otimes\Imat}
\end{align*}
Thus, the final Hessian is given by:
\begin{align*}
    \frac{\dd \phi(\xvec,\Amat)}{\dd \xvec^2} &= \Amat + \Amat^T\\
    \frac{\dd \phi(\xvec,\Amat)}{\dd \xvec^T\,\dd\vvec{\Amat}}  &= \pare{\xvec^T\otimes\Imat} + \pare{\Imat \otimes \xvec^T}  \\
    \frac{\dd \phi(\xvec,\Amat)}{\dd\vvecT{\Amat}\,\dd \xvec} &= \pare{\xvec \otimes \Imat} + \pare{\Imat \otimes \xvec}\\
    \frac{\dd \phi(\xvec,\Amat)}{\dd\vvecT{\Amat}\,\dd\vvec{\Amat}} &= \zeromat
\end{align*}
Since both $\xvec \in \realN, \vvec\Amat \in \realNsquare$ , then submatrices $\frac{\dd \phi(\xvec,\Amat)}{\dd \xvec^T\,\dd\vvec{\Amat}}$ and $ \frac{\dd \phi(\xvec,\Amat)}{\dd\vvecT{\Amat}\,\dd \xvec} $ cannot be symmetric, while others are, since they are the Hessians corresponding to functions with arguments being $\Amat$ and $\xvec$. However, since it is the full Hessian what needs to be symmetric, and since this Hessian has been developed by decomposing into a $2\times 2$ block matrix. Then because $\blockTranspose$. We have:
\begin{align*}
    \braT{\frac{\dd \phi(\xvec,\Amat)}{\dd \xvec^2}} &= \Amat + \Amat^T = \frac{\dd \phi(\xvec,\Amat)}{\dd \xvec^2} && \text{using } \sumTranspose\\
    \braT{\frac{\dd \phi(\xvec,\Amat)}{\dd \xvec^T\,\dd\vvec{\Amat}}}  &= \pare{\xvec\otimes\Imat} + \pare{\Imat \otimes \xvec}  =  \frac{\dd \phi(\xvec,\Amat)}{\dd\vvecT{\Amat}\,\dd \xvec} && \text{using } \sumTranspose,\kronTranspose \\
    \braT{\frac{\dd \phi(\xvec,\Amat)}{\dd\vvecT{\Amat}\,\dd \xvec}} &= \pare{\xvec^T \otimes \Imat} + \pare{\Imat \otimes \xvec^T} = \frac{\dd \phi(\xvec,\Amat)}{\dd \xvec^T\,\dd\vvec{\Amat}} && \text{using } \sumTranspose,\kronTranspose \\
    \braT{\frac{\dd \phi(\xvec,\Amat)}{\dd\vvecT{\Amat}\,\dd\vvec{\Amat}}} &= \zeromat = \frac{\dd \phi(\xvec,\Amat)}{\dd\vvecT{\Amat}\,\dd\vvec{\Amat}}
\end{align*}
We can conclude that the full Hessian is symmetric. This sanity check will be used later and will allow me to showcase some particularities I have found when performing some derivations. 
\section{Useful math identities}
Here is a list of the mathematical identities used in this work. Most of them are proven and developed by \cite{magnus99}. Some of them might also be found in \cite{minka,matcookbook}. Those that are not, are proven in this document. I  also extend one of the theorems from \cite{magnus99} to accommodate this work.

\begin{align}
    %% VEC OPERATOR
    \vvecB{\avec^T} &= \vvec{\avec} = \avec \label{vec_of_a_vec}\\
    \vvecB{\Amat\Xmat\Bmat} &= \pare{\Bmat^T\otimes\Amat}\vvec{\Xmat} \label{vecmat2kron}\\
    \vvecB{\Amat\Bmat\dvec} &= \bra{\dvec^T \otimes \Amat } \vvec \Bmat = \bra{\Amat \otimes \dvec^T} \vvec{\Bmat^T}\\
    \vvecB{\Amat+\Bmat} &= \vvec \Amat+\vvec\Bmat;\,\,\, \vvecBT{\Amat+\Bmat} = \vvecBT{\Amat}+\vvecBT{\Bmat}  \label{vecsum}\\
    \vvec\Amat &= \Dn\vvech\Amat  \label{vec2vech}\\
    %
    %
    %
    % TRACE OPERATOR
    \trB{\avec^T\bvec} &= \avec^T\bvec=\vvecT{\avec}\vvec{\bvec}  \label{tr_dotprod_2vec}\\
    \trB{\Amat^T\Bmat} &= \vvecT{\Amat}\vvec{\Bmat}  \label{tr_matprod_2vec}\\
    \trB{\Amat\Bmat\cvec\dvec^T} &= \vvecBT{\Amat^T}\pare{\dvec\otimes \Bmat} \cvec \label{tr_ABcdT_2vec}\\
    \trB{\Amat \Bmat \Cmat \Dmat} &= \vvecT{\Bmat^T} \pare{\Amat^T \otimes \Cmat} \vvec{\Dmat} \label{tr_ABCD_2vec}\\
    %
    %
    %\ddB{\vvec{\Xmat}} &= \vvecB{\ddsymbol \Xmat} \\
    %\trB{\Amat^T\Bmat} &= \vvecT{\Amat}\vvec{\Bmat} \\
    %\trB{\Amat}\trB{\Bmat} &= \trB{\Amat \otimes \Bmat}\\
    %\vvecB{\avec\bvec^T} &= \bvec \otimes \avec \\
    %\vvecB{\Amat\cvec\dvec^T\Bmat} &= \pare{\Bmat^T\otimes \Amat}\vvecB{\cvec\dvec^T}\\
    %% Expressions for differentials and Hessian
    \dd^2\phi(\Xmat)&= \trB{\Amat\braT{\dd\Xmat}\Cmat\dd\Xmat} \Rightarrow \frac{\dd^2\phi(\Xmat)}{\dd\vvecT{\Xmat}\dd\vvec{\Xmat}} = \frac{1}{2}\pare{\Amat^T\otimes \Cmat+\Amat\otimes \Cmat^T} \label{tr_dXdX_2Hess}\\
    \dd^2\phi(\Xmat,\Ymat)&= \trB{\Amat\braT{\dd\Ymat}\Cmat\dd\Xmat} \Rightarrow \frac{\dd^2\phi(\Xmat,\Ymat)}{\dd\vvecT{\Ymat}\dd\vvec{\Xmat}} = \frac{1}{2}\pare{\Amat^T\otimes \Cmat+\Amat\otimes \Cmat^T}\iff \dim\Amat = \dim\Cmat = N^2 \label{tr_dYdX_2Hess}\\
    %
    %
    %% DIFERENTIAL RULES
    \dd \Bdd{ \Xmat } &= 0 \label{d2X}\\
    \dd \Amat &= 0 \label{dconst}\\
    \ddB{\Amat\Xmat} &= \Amat\dd\Xmat \label{dAX}\\
    \ddB{\Xmat^T} &= \braT{\dd\Xmat} \label{dXT} \\
    \ddB{\Xmat + \Ymat} &= \dd\Xmat + \dd\Ymat \label{dsum}\\
    \ddB{\tr \Xmat} &= \trB{\dd\Xmat} \label{dTR}\\
    \ddB{\Xmat\Ymat} &= \Xmat\dd\Ymat + \Bdd{\Xmat}\Ymat \label{dprod}\\
    \ddB{\Xmat^{-1}} &= -\Xmat^{-1} \dd \Xmat \Xmat^{-1} \label{dinv}\\
    \dd \left|\Xmat\right|  &= \left|\Xmat\right|\trB{\Xmat^{-1}\dd\Xmat} \label{ddet}\\
    \dd \log \Xmat &= \frac{1}{\Xmat}\dd\Xmat \label{dlog}\\
    \ddB{\vvec{\Xmat}} &= \vvec{\dd \Xmat} \label{dvec}\\
    \ddB{\vvech{\Xmat}} &= \vvech{\dd \Xmat}  \label{dvech} 
\end{align}    

Many other well-known matrix identities are used throughout the derivations. All of them can be found in \cite{matcookbook}. While performing derivations, I will clearly refer to its usage.

\subsection{Proofs}
Proofs for some identities I have not found in the literature yet are needed for this work. 

\begin{proposition}[]
$\trB{\Amat\Bmat\cvec\dvec^T} = \vvecBT{\Amat^T}\pare{\dvec\otimes \Bmat} \cvec$
\end{proposition}
Because $\circularTR$ and noting that $\dvec^T\Amat\Bmat$ is a row vector.
\begin{align*}
    \trB{\Amat\Bmat\cvec\dvec^T} &= \trB{\dvec^T\Amat\Bmat\cvec} && \\
    &= \vvecT{\pare{\braT{\dvec^T\Amat\Bmat}}}\vvec{\cvec} && \text{using \eqns}  \ref{tr_dotprod_2vec}, \ref{vec_of_a_vec}\\
    &= \vvecT{\pare{\Bmat^T\Amat^T\dvec}}\vvec{\cvec} && \text{using} \prodTranspose\\
    &= \braT{\pare{\dvec^T \otimes \Bmat^T}\vvecB{\Amat^T}}\vvec{\cvec} && \text{using \eqn}  \ref{vecmat2kron}\\
    &=\vvecBT{\Amat^T} \pare{\dvec \otimes \Bmat} \vvec{\cvec} && \text{using} \prodTranspose;\, \kronTranspose\\
    &=\vvecBT{\Amat^T} \pare{\dvec \otimes \Bmat} \cvec && \text{using \eqn} \ref{vec_of_a_vec} 
\end{align*}

In case $\Amat$ is symmetric $\symetricTranspose$ the result is:
\begin{align*}
&\vvecT{\Amat} \pare{\dvec \otimes \Bmat} \cvec
\end{align*}
This result will be very important to obtain a submatrix from the Hessian. It is used in \citep{minka} to go from \eqn $123$ to \eqn $124$, yet the identity is not explicitly derived.

\begin{proposition}[This is an extension from theorem 10.1 in \cite{magnus99}]
\begin{align*}
     \dd^2\phi(\Xmat,\Ymat)= \trB{\Amat\braT{\dd\Ymat}\Cmat\dd\Xmat} \Rightarrow \frac{\dd^2\phi(\Xmat,\Ymat)}{\dd\vvecT{\Ymat}\dd\vvec{\Xmat}} = \frac{1}{2}\pare{\Amat^T\otimes \Cmat+\Amat\otimes \Cmat^T}\iff \dim\Amat = \dim\Cmat = N\times N
\end{align*}
\end{proposition}
Using \eqn \ref{tr_ABCD_2vec} we can write:
\begin{align*}
\trB{\Amat\braT{\dd\Ymat}\Cmat\dd\Xmat}  &= \vvecT{\dd\Ymat} \pare{\Amat^T \otimes \Cmat}  \vvec{\dd\Xmat}
\end{align*}
Then, using the second identification theorem, we have:
\begin{align*}
    \Bmat =  \pare{\Amat^T \otimes \Cmat}
\end{align*}
Since we require this differential to be symmetric, we apply (theorem 6.6 in \cite{magnus99}):
\begin{align*}
    \frac{\dd^2\phi(\Xmat,\Ymat)}{\dd\vvecT{\Ymat}\dd\vvec{\Xmat}} = \Bmat + \Bmat^T = \frac{1}{2}\pare{\Amat^T\otimes \Cmat+\Amat\otimes \Cmat^T}
\end{align*}
In order to perform this computation, we need $\dim\Amat = \dim\Cmat = N\times N$; otherwise computation is unfeasible. This implies that $\dim\Xmat = \dim\Ymat = N \times N$ otherwise computation of $\trB{\Amat\braT{\dd\Ymat}\Cmat\dd\Xmat}$  is also unfeasible. This is easily seen by noting that any other dimensionality will result in either a computation that cannot be performed or a non-square matrix over which the trace is not defined.

This result is particularly useful to obtain some partial derivatives when computing the Hessian.
\section{Jacobian \KLD}
Consider two $N$-dimensional multivariate Gaussian distributions:
\begin{align*}
    p(\xvec) &= \mathcal{N}\left(\xvec\mid\wvec,\Vmat\right)\\
    q(\xvec) &= \mathcal{N}\left(\xvec\mid\mvec,\Smat\right)
\end{align*}
then:
\begin{align*}
    \KLD[q(\xvec)\mid\mid p(\xvec)] &= \frac{1}{2}\bra{\log\mid\Vmat\mid - \log \mid\Smat\mid -N +\trB{\Vmatinv\Smat}+(\mvec-\wvec)^T\Vmatinv(\mvec-\wvec)}
\end{align*}
with $\mvec,\wvec \in \realN$ and $\Vmat,\Smat \in \mathbb{S}^N_{++} \subset \realNN$, \ie the subspace of $N\times N$ positive definitive matrices. We know that covariance matrices are symmetric, which implies $\Smat=\Smatt,\Vmat=\Vmatt$.

As noted previously, the Jacobian matrix can be partitioned into the differentials \wrt the different variables, and then summing up the different contributions. We will use this fact to compute the full Jacobian by taking differentials \wrt the different variables separately.

\subsection{\texorpdfstring{Jacobian \wrt $\mvec$}{Jacobian m}}

Here we compute the Jacobian \wrt a vector. Thus, we operate on $\phi: \realN \to \real$. Let $\avec=(\mvec-\wvec)$, we can obtain the first order differential \wrt $\mvec$ by applying the chain rule. Thus:
\begin{align*}
    \dd_{\avec} \KLD &= \ddB{ \frac{1}{2}\avec^T\Vmatinv\avec} && \text{using \eqns } \ref{dsum},\ref{dconst}\\
    &= \frac{1}{2}\bra{\ddB{\avec^T}\Vmatinv\avec +\avec^T\Vmatinv\dd\avec} && \text{using \eqns } \ref{dprod},\ref{dAX} \\
    &= \frac{1}{2}\bra{\avec^T\Vmatinvt\dd\avec +\avec^T\Vmatinv\dd\avec} && \text{using \eqn } \ref{dXT},\, \quadraticScalar,\scalarTranspose,\prodTranspose\\ 
    &= \avec^T\Vmatinv\dd\avec && \text{using } \transposeInverse, \symetricTranspose
\end{align*}
and 
\begin{align*}
    \dd_{\mvec}(\mvec - \wvec) = \Imat\dd\mvec && \text{using \eqns } \ref{dsum},\ref{dconst}, \matprodIdentity
\end{align*}
Applying the chain rule of the first-order differential, we have:
\cbox{
\textit{\textbf{Differential}}:
\begin{align*}
    \dd_\mvec \KLD = (\mvec-\wvec)^T\Vmatinv \Imat\dd\mvec
\end{align*}
}
from which we identify the Jacobian:
\cbox{
\textit{\textbf{Jacobian}}:
\begin{align*}
    \Jacobian{\KLD,\mvec} = (\mvec-\wvec)^T\Vmatinv  \in \real^{1\times N}
\end{align*}
}
we can verify that the Jacobian lives in $\real^{1\times N}$ as expected from the table of the canonical forms.
\subsection{\texorpdfstring{Jacobian \wrt $\wvec$}{Jacobian w}}
This Jacobian can be computed in a similar way to the previous one, which implies operating on $\phi: \realN \to \real$. Since the dependence on the \KLD is only through the quadratic form. We can apply the same chain rule and only compute the second term. Thus:
\begin{align*}
    \dd_{\avec} \KLD &= \ddB{ \frac{1}{2}\avec^T\Vmatinv\avec}  && \text{using \eqns } \ref{dsum},\ref{dconst} \\
    &= \frac{1}{2}\bra{\ddB{\avec^T}\Vmatinv\avec +\avec^T\Vmatinv\dd\avec} && \text{using \eqns } \ref{dprod},\ref{dAX} \\
    &= \frac{1}{2}\bra{\avec^T\Vmatinvt\dd\avec +\avec^T\Vmatinv\dd\avec} && \text{using \eqn } \ref{dXT},\, \quadraticScalar,\scalarTranspose,\prodTranspose\\ 
    &= \avec^T\Vmatinv\dd\avec && \text{using } \transposeInverse, \symetricTranspose
\end{align*}
and 
\begin{align*}
    \dd_{\wvec}(\mvec - \wvec) = -\Imat\dd\wvec && \text{using \eqns } \ref{dsum},\ref{dconst}, \matprodIdentity
\end{align*}
Which gives a first-order differential:
\cbox{
\textit{\textbf{Differential}}:
\begin{align*}
    \dd_\wvec \KLD = -(\mvec-\wvec)^T\Vmatinv \Imat\dd\wvec
\end{align*}
}
from which we identify the Jacobian:
\cbox{
\textit{\textbf{Jacobian}}:
\begin{align*}
    \Jacobian{\KLD,\wvec} = (\wvec-\mvec)^T\Vmatinv  \in \real^{1\times N}
\end{align*}
}
\subsection{\texorpdfstring{Jacobian \wrt $\Smat$}{Jacobian S}}
Here we compute the Jacobian \wrt a symmetric matrix. Thus, we now operate on $\phi: \mathbb{S}^N_{++} \to \real$, and require the $\vecsymbol$ operator to express the differential in its canonical form.
\begin{align*}
    \dd_\Smat \KLD &= \frac{1}{2}\dd\bra{-\log \mid\Smat \mid + \trB{\Vmatinv\Smat}}  && \text{using \eqns } \ref{dsum},\ref{dconst}
\end{align*}
Since the differential of the sum is the sum of the differentials, we can compute the terms independently:
\begin{align*}
    \dd_\Smat\bra{-\log \mid\Smat \mid } &= -\ddB{\log \mid\Smat \mid}  &&\\
    &= -\frac{1}{\mid\Smat \mid}\dd \mid\Smat \mid && \text{using \eqn}\ref{dlog}\\
    &= -\cancel{\frac{1}{\mid\Smat \mid}}\cancel{\left|\Smat\right|}\trB{\Smatinv\dd\Smat} && \text{using \eqn}\ref{ddet}\\
    &= -\vvecBT{\Smatinvt}\dd\vvec{\Smat} && \text{using \eqns}\ref{tr_matprod_2vec},\ref{dvec}\\
    &= -\vvecT{\Smatinv}\dd\vvec{\Smat} && \text{using } \transposeInverse,\symetricTranspose
\end{align*}
and
\begin{align*}
    \dd_\Smat \trB{\Vmatinv\Smat} &= \trB{\Vmatinv\dd\Smat} && \text{using \eqns}\ref{dTR},\ref{dAX}\\
    &= \vvecT{\Vmatinv}\dd\vvec{\Smat} && \text{using \eqns}\ref{tr_matprod_2vec},\ref{dvec}, \transposeInverse,\symetricTranspose
\end{align*}
Thus we have:
\begin{align*}
     \dd_\Smat \KLD &= \frac{1}{2}\dd\bra{-\log \mid\Smat \mid + \trB{\Vmatinv\Smat}} && \\
     &= \frac{1}{2}\bra{\vvecT{\Vmatinv}\dd\vvec{\Smat} -\vvecT{\Smatinv}\dd\vvec{\Smat}} && \\
     &= \frac{1}{2}\vvecBT{\Vmatinv-\Smatinv}\dd\vvec{\Smat} &&  \text{using \eqn} \ref{vecsum}
\end{align*}
\cbox{
\textit{\textbf{Differential}}:
\begin{align*}
    \dd_{\vvec\Smat} \KLD &= \frac{1}{2}\vvecBT{\Vmatinv-\Smatinv}\dd\vvec{\Smat}
\end{align*}
}
from where the Jacobian is given by:
\cbox{
\textit{\textbf{Jacobian}}:
\begin{align*}
    \Jacobian{\KLD,\vvec{\Smat}} = \frac{1}{2}\vvecBT{\Vmatinv-\Smatinv}  \in \realNsquarerow
\end{align*}
}
If we only consider the unique elements of $\Smat$, since is a symmetric matrix, then we have:
\begin{align*}
     \dd_\Smat \KLD &= \frac{1}{2}\vvecBT{\Vmatinv-\Smatinv}\dd\vvec{\Smat} && \\
     &= \frac{1}{2}\vvecBT{\Vmatinv-\Smatinv}\Dn\dd\vvech{\Smat} &&  \text{using \eqns} \ref{vec2vech},\ref{dAX}
\end{align*}
\cbox{
\textit{\textbf{Differential}}:
\begin{align*}
     \dd_{\vvech\Smat} \KLD &= \frac{1}{2}\vvecBT{\Vmatinv-\Smatinv}\Dn\dd\vvech{\Smat} 
\end{align*}
}
thus:
\cbox{
\textit{\textbf{Jacobian}}:
\begin{align*}
    \Jacobian{\KLD,\vvech{\Smat}} = \frac{1}{2}\vvecBT{\Vmatinv-\Smatinv}\Dn  \in \real^{1\times \nicefrac{1}{2}(N^2 + N)}
\end{align*}
}
\subsection{\texorpdfstring{Jacobian \wrt $\Vmat$}{Jacobian V}}
We follow previous section operating on functions $\phi: \mathbb{S}^N_{++} \to \real$, requiring again the $\vecsymbol$ operator. The first-order differential is computed by operating on:
\begin{align*}
    \dd_\Vmat \KLD = \frac{1}{2}\dd\bra{\log \mid\Vmat \mid + \trB{\Vmatinv\Smat} + \avec^T\Vmatinv\avec};\,\, \avec = (\mvec-\wvec)  && \text{using \eqns } \ref{dsum},\ref{dconst}
\end{align*}
Operating on the terms separately, due to the linearity of differential, we have:
\begin{align*}
    \dd_\Vmat\bra{\log \mid\Vmat \mid } &= \ddB{\log \mid\Vmat \mid}  && \\
    &= \frac{1}{\mid\Vmat\mid}\dd \mid\Vmat \mid && \text{using \eqn} \ref{dlog}\\
    &= \cancel{\frac{1}{\mid\Vmat\mid}}\cancel{\left|\Vmat\right|}\trB{\Vmatinv\dd\Vmat} && \text{using \eqn} \ref{ddet}\\
    &= \vvecBT{\Vmatinvt}\dd\vvec{\Vmat} && \text{using \eqns} \ref{tr_matprod_2vec},\ref{dvec}\\
    &= \vvecT{\Vmatinv}\dd\vvec{\Vmat} && \text{using } \transposeInverse,\symetricTranspose
\end{align*}
and
\begin{align*}      
\dd_\Vmat\bra{\trB{\Vmatinv\Smat}} &= \trB{\dd\Vmatinv\Smat}  && \text{using \eqns} \ref{dTR},\ref{dAX}\\
    &= -\trB{\Vmatinv\dd\Vmat\Vmatinv\Smat}  && \text{using \eqn} \ref{dinv}\\
    &= -\trB{\Vmatinv\Smat\Vmatinv\dd\Vmat} && \text{using } \circularTR \\
    &= -\vvecBT{\braT{\Vmatinv\Smat\Vmatinv}}\vvec{\dd\Vmat} && \text{using \eqn } \ref{tr_matprod_2vec} \\
    &= -\vvecBT{\Vmatinv\Smat\Vmatinv}\dd\vvec{\Vmat} && \text{using \eqn} \ref{dvec},\prodTranspose,\transposeInverse,\symetricTranspose
\end{align*}
Finally, we note that the differential of the quadratic form can be obtained in a similar way to the last one, by noting that $\avec\avec^T$ is an $N\times N$ symmetric matrix as $\Smat$. Since:
\begin{align*}
    \pareT{\avec\avec^T} &= \avec\avec^T && \text{using }\prodTranspose
\end{align*}
then:
\begin{align*}
    \dd_\Vmat \bra{\avec^T\Vmatinv\avec} &= \dd\bra{\trB{\Vmatinv\avec\avec^T}}  && \text{using } \circularTR\\
    &= -\vvecBT{\Vmatinv\avec\avec^T\Vmatinv}\dd\vvec{\Vmat} && \text{using the previous derivation for } \dd_\Vmat\bra{\trB{\Vmatinv\Smat}}
\end{align*}
Thus we have:
\begin{align*}
      \dd_\Vmat \KLD &= \frac{1}{2}\pare{\vvecT{\Vmatinv}\dd\vvec{\Vmat}-\vvecBT{\Vmatinv\Smat\Vmatinv}\dd\vvec{\Vmat}-\vvecBT{\Vmatinv\avec\avec^T\Vmatinv}\dd\vvec{\Vmat}}\\
      &= \frac{1}{2}\pare{\vvecT{\Vmatinv}-\vvecBT{\Vmatinv\Smat\Vmatinv}-\vvecBT{\Vmatinv\avec\avec^T\Vmatinv}}\dd\vvec{\Vmat}\\
      &= \frac{1}{2}\vvecBT{\Vmatinv-\Vmatinv\Smat\Vmatinv-\Vmatinv\avec\avec^T\Vmatinv}\dd\vvec{\Vmat} && \text{using \eqn } \ref{vecsum}\\
      &= \frac{1}{2}\vvecBT{\Vmatinv-\Vmatinv\pare{\Smat+\avec\avec^T}\Vmatinv}\dd\vvec{\Vmat}
\end{align*}
\cbox{
\textit{\textbf{Differential}}:
\begin{align*}
    \dd_{\vvec\Vmat} \KLD &=\frac{1}{2}\vvecBT{\Vmatinv-\Vmatinv\pare{\Smat+(\mvec-\wvec)(\mvec-\wvec)^T}\Vmatinv}\dd\vvec{\Vmat}
\end{align*}
}
and so the Jacobian is given by:
\cbox{
\textit{\textbf{Jacobian}}:
\begin{align*}
     \Jacobian{\KLD,\vvec{}\Vmat} = \frac{1}{2}\vvecBT{\Vmatinv-\Vmatinv\pare{\Smat+(\mvec-\wvec)(\mvec-\wvec)^T}\Vmatinv}  \in \realNsquarerow
\end{align*}
}
Considering only the unique elements of $\Vmat$ we have:
\begin{align*}
     \dd_\Vmat \KLD &= \frac{1}{2}\vvecBT{\Vmatinv-\Vmatinv\pare{\Smat+\avec\avec^T}\Vmatinv}\dd\vvec{\Vmat} && \\
     &= \frac{1}{2}\vvecBT{\Vmatinv-\Vmatinv\pare{\Smat+(\mvec-\wvec)(\mvec-\wvec)^T}\Vmatinv}\Dn\dd\vvech{\Vmat} &&  \text{using \eqns } \ref{vec2vech}, \ref{dAX}
\end{align*}
\cbox{
\textit{\textbf{Differential}}:
\begin{align*}
     \dd_{\vvech\Vmat} \KLD &= \frac{1}{2}\vvecBT{\Vmatinv-\Vmatinv\pare{\Smat+(\mvec-\wvec)(\mvec-\wvec)^T}\Vmatinv}\Dn\dd\vvech{\Vmat}
\end{align*}
}
thus:
\cbox{
\textit{\textbf{Jacobian}}:
\begin{align*}
    \Jacobian{\KLD,\vvech{\Vmat}} =\frac{1}{2}\vvecBT{\Vmatinv-\Vmatinv\pare{\Smat+(\mvec-\wvec)(\mvec-\wvec)^T}\Vmatinv}\Dn \ \in \real^{1\times \nicefrac{1}{2}(N^2 + N)}
\end{align*}
}
\section{Hessian \KLD}
We will use the Hessian to showcase some of the different options we have to compute it, although the canonical form of the second-order differential will always be the preferred option.

The full Hessian is a matrix with the following partial derivatives:
\begin{align*}
    \Hessian{\KLD} = \begin{bmatrix}
\frac{\dd \KLD}{\dd \mvec^2} & \frac{\dd \KLD}{\dd \mvec^T\,\dd\wvec} & \frac{\dd \KLD}{\dd \mvec^T\,\dd\vvec{\Smat}} & \frac{\dd \KLD}{\dd \mvec^T\,\dd\vvec{\Vmat}} \\\\
\frac{\dd \KLD}{\dd \wvec^T\,\dd \mvec} & \frac{\dd \KLD}{\dd \wvec^2} & \frac{\dd \KLD}{\dd \wvec^T\,\dd\vvec{\Smat}} & \frac{\dd \KLD}{\dd \wvec^T\,\dd\vvec{\Vmat}} \\\\
\frac{\dd \KLD}{\dd\vvecT{\Smat}\,\dd \mvec} & \frac{\dd \KLD}{\dd\vvecT{\Smat}\,\dd\wvec} & \frac{\dd \KLD}{\dd\vvec{\Smat}^2} & \frac{\dd \KLD}{\dd\vvecT{\Smat}\,\dd\vvec{\Vmat}} \\\\
\frac{\dd \KLD}{\dd\vvecT{\Vmat}\,\dd \mvec} & \frac{\dd \KLD}{\dd\vvecT{\Vmat}\,\dd\wvec} & \frac{\dd \KLD}{\dd\vvecT{\Vmat}\,\dd\vvec{\Smat}} & \frac{\dd \KLD}{\dd\vvec{\Vmat}^2}
\end{bmatrix}
\end{align*}
\subsection{\texorpdfstring{Partial derivative $\frac{\dd \KLD}{\dd \mvec^T\,\dd\mvec}$}{Partial dm dm}}
Here we deal with $\phi: \realN \times \realN \to \real$. 
\subsubsection{From second-order differential}
First, we obtain the Hessian by computing the second-order differential. To do so, we compute the differential of the first-order differential. 

The first-order differential is:
\begin{align*}
     \dd_\mvec \KLD = (\mvec-\wvec)^T\Vmatinv \dd\mvec
\end{align*}
So the second-order differential is given by:
\begin{align*}
     \dd^2_\mvec \KLD &= \dd_\mvec\bra{(\mvec-\wvec)^T\Vmatinv \dd\mvec} && \\
     &=\bra{\ddB{(\mvec-\wvec)^T}\Vmatinv \dd\mvec} + (\mvec-\wvec)^T\Vmatinv \cancelto{0}{\ddB{\dd\mvec}} && \text{using \eqns } \ref{dprod},\ref{dAX},\ref{d2X} \\
     &= \Tdd{(\mvec-\wvec)}\Vmatinv \dd\mvec && \text{using \eqn } \ref{dXT}\\
     &= \Tdd{\mvec}\Vmatinv \dd\mvec && \text{using \eqns } \ref{dsum},\ref{dconst}
\end{align*}
\cbox{
\textit{\textbf{Differential}}
\begin{align*}
    \dd^2_\mvec \KLD = \Tdd{\mvec}\Vmatinv \dd\mvec
\end{align*}
}
from where:
\cbox{
\textit{\textbf{Hessian}}
\begin{align*}
   \frac{\dd \KLD}{\dd \mvec^T\,\dd\mvec} = \Vmatinv \in \realNN
\end{align*}
}
which is symmetric, so it is directly identified as the Hessian submatrix. Note that if we fix all variables but $\mvec$ this would be directly the Hessian of the function.
\subsubsection{From the first-order differential of the transposed Jacobian}
We have:
\begin{align*}
    \Jacobian{\KLD,\mvec} = (\mvec-\wvec)^T\Vmatinv
\end{align*}
Thus:
\begin{align*}
    \dd_\mvec  \Jacobian{\KLD,\mvec}^T &= \ddB{\braT{(\mvec-\wvec)^T\Vmatinv}} \\
    &= \ddB{\Vmatinv(\mvec-\wvec)} && \text{using } \prodTranspose, \transposeInverse, \symetricTranspose, \\
    &= \Vmatinv\dd\mvec && \text{using \eqns } \ref{dAX},\ref{dsum},\ref{dconst}
\end{align*}
Since for this differential, we identify its corresponding Jacobian as $\Vmatinv$ this implies that:
\begin{align*}
    \frac{\dd \KLD}{\dd \mvec^T\,\dd\mvec} =  \Vmatinv,
\end{align*}
as expected.
\subsubsection{By direct computation}
We can directly derive the transposed Jacobian, instead of obtaining its first differential, to derive the Hessian. Thus:
\begin{align*}
    \frac{\dd \Jacobian{\KLD,\mvec}^T}{\ddT{\mvec}}   &= \frac{\dd}{\ddT{\mvec}}\braT{(\mvec-\wvec)^T\Vmatinv} && \\
    &= \frac{\dd}{\ddT{\mvec}}\Vmatinv(\mvec-\wvec) && \text{using \eqn } \prodTranspose,\transposeInverse,\symetricTranspose\\
    &= \Vmatinv\cancelto{\Imat}{\frac{\dd}{\ddT{\mvec}}\mvec} && \text{using linearity of derivative and computing partial derivatives explicitly.} \\
    &= \Vmatinv
\end{align*}
\subsection{\texorpdfstring{Partial derivative: $\frac{\dd \KLD}{\dd \mvec^T\,\dd\wvec}$}{Partial dm dw}}
Here we deal with $\phi:\realN \times \realN \to \real$. 

The first differential \wrt $\wvec$ is:
\begin{align*}
     \dd_\wvec \KLD = -(\mvec-\wvec)^T\Vmatinv \dd\wvec&& 
\end{align*}
The second differential \wrt $\mvec$ is:
\begin{align*}
     \dd_\mvec \dd_\wvec \KLD &=  \dd_\mvec\bra{-(\mvec-\wvec)^T\Vmatinv \dd\wvec}&& \\
     &= \dd_\mvec\bra{-\mvec^T\Vmatinv \dd\wvec}+ \dd_\mvec\bra{\cancelto{0}{\wvec^T\Vmatinv \dd\wvec}}&& \text{using \eqns } \ref{dconst},\ref{dsum}\\
    &= \braT{\dd\mvec}-\Vmatinv \dd\wvec&& \text{using \eqns } \ref{dXT},\ref{dAX}
\end{align*}
\cbox{
\textit{\textbf{Differential}}
\begin{align*}
 \dd_\mvec\dd_\wvec \KLD = \braT{\dd\mvec}-\Vmatinv \dd\wvec
\end{align*}
}
So we identify:
\cbox{
\textit{\textbf{Hessian}}
\begin{align*}
    \frac{\dd \KLD}{\dd \mvec^T\,\dd\wvec} = -\Vmatinv  \in \realNN
\end{align*}
}
Note that, while being symmetric, this is not identified as the Hessian from any function. As we'll see later there are some partial derivatives of this type which are not symmetric.
\subsection{\texorpdfstring{Partial derivative: $\frac{\dd \KLD}{\dd\mvec^T\,\dd\vvec{\Smat}}$}{Partial dm dS}}
Here we deal with $\phi: \realN \times \mathbb{S}^N_{++} \to \real$.

Because the first differential is:
\begin{align*}
    \dd_\Smat\KLD = \frac{1}{2}\vvecBT{\Vmatinv-\Smatinv}\dd\vvec{\Smat}
\end{align*}
The second differential \wrt $\mvec$ is:
\begin{align*}
    \dd_\mvec\dd_\Smat \KLD &= \dd_\mvec\bra{\frac{1}{2}\vvecBT{\Vmatinv-\Smatinv}\dd\vvec{\Smat}}\\
    &= \zeromat && \text{since none of the terms depends on } \mvec
\end{align*}
\cbox{
\textit{\textbf{Differential}}
\begin{align*}
    \dd_\mvec \dd_{\vvec{\Smat}} \KLD &= \braT{\dd \mvec} \zeromat \dd\vvec{\Smat}
\end{align*}
}
So we identify:
\cbox{
\textit{\textbf{Hessian}}
\begin{align*}
    \frac{\dd \KLD}{\dd\mvec^T\,\dd\vvec{\Smat}} = \zeromat \in \realNNsquare
\end{align*}
}
The resulting partial derivative does not represent the Hessian of any function, which is why it is not a symmetric matrix.

Considering only the unique elements of $\Smat$, we have:

\cbox{
\textit{\textbf{Differential}}
\begin{align*}
    \dd_\mvec \dd_{\vvech{\Smat}} \KLD &= \braT{\dd \mvec} \zeromat \Dn \dd\vvech{\Smat} && \text{using \eqns } \ref{vec2vech}, \ref{dAX}
\end{align*}
}
\cbox{
\textit{\textbf{Hessian}}
\begin{align*}
    \frac{\dd \KLD}{\dd\mvec^T\,\dd\vvech{\Smat}} = \zeromat \Dn \in \realNNDup
\end{align*}
}
\subsection{\texorpdfstring{Partial derivative $\frac{\dd \KLD}{\dd\mvec^T \dd \vvec\Vmat}$}{Partial dm dV}}\label{dmdv_derivation}
Here we deal with $\phi: \realN \times \mathbb{S}^N_{++} \to \real$.

Because:
\begin{align*}
        \dd_\Vmat \KLD &= \frac{1}{2}\vvecBT{\Vmatinv-\Vmatinv\pare{\Smat+(\mvec-\wvec)(\mvec-\wvec)^T}\Vmatinv}\dd\vvec{\Vmat} && \\
       &=\frac{1}{2}\bra{\trB{\Vmatinv\dd\Vmat}-\trB{\Vmatinv\Smat\Vmatinv\dd\Vmat}-\trB{\Vmatinv\avec\avec^T\Vmatinv\dd\Vmat}};\, \avec = (\mvec-\wvec) && \text{using \eqns } \ref{tr_matprod_2vec},\ref{vecsum},\ref{dvec}\\
       & && \prodTranspose,\\
       & && \transposeInverse,\symetricTranspose
\end{align*}
then:
\begin{align*}
      \dd_\mvec\dd_\Vmat \KLD &= \dd_\mvec\Big[\frac{1}{2}\Big(\cancelto{0}{\trB{\Vmatinv\dd\Vmat}}\\
      & -\cancelto{0}{\trB{\Vmatinv\Smat\Vmatinv\dd\Vmat}}-\trB{\Vmatinv\avec\avec^T\Vmatinv\dd\Vmat}\Big)\Big] && \text{using \eqns } \ref{dsum},\ref{dconst}\\
      &=-\frac{1}{2}\dd_\mvec\bra{\trB{\avec^T\Vmatinv\dd\Vmat\Vmatinv\avec}}&& \text{using } \circularTR\\
      &=-\frac{1}{2}\trB{\ddB{\avec^T}\Vmatinv\dd\Vmat\Vmatinv\avec + \avec^T\Vmatinv\dd\Vmat\Vmatinv\dd\avec}&& \text{using \eqns } \ref{dprod},\ref{dTR},\ref{dAX}\\
      &=-\frac{1}{2}\trB{\avec^T\Vmatinv\dd\Vmat\Vmatinv\dd\avec + \avec^T\Vmatinv\dd\Vmat\Vmatinv\dd\avec}&& \text{using  \eqn} \ref{dXT},\quadraticScalar,\scalarTranspose,\\
      & && \prodTranspose,\transposeInverse,\symetricTranspose\\
      &=-\trB{(\mvec-\wvec)^T\Vmatinv\dd\Vmat\Vmatinv\dd\mvec}&& \text{using \eqns } \ref{dconst},\ref{dsum}\\
      &=-\trB{\braT{\dd\mvec}\Vmatinv\dd\Vmat\Vmatinv(\mvec-\wvec)}&& \text{using } \quadraticScalar,\scalarTranspose,\\
      & && \prodTranspose,\transposeInverse,\symetricTranspose\\
      &=-\trB{\Vmatinv(\mvec-\wvec)\braT{\dd\mvec}\Vmatinv\dd\Vmat}&& \text{using } \circularTR \\
      &=-\vvecBT{\dd\mvec}\pare{\braT{\Vmatinv(\mvec-\wvec)}\otimes \Vmatinv}\vvec{\dd\Vmat}&& \text{using \eqn } \ref{tr_ABCD_2vec} \\
      &=\braT{\dd\mvec}-\pare{\bra{(\mvec-\wvec)^T\Vmatinv}\otimes \Vmatinv}\dd\vvec{\Vmat}&& \text{using \eqns } \ref{dvec},\ref{vec_of_a_vec}, \prodTranspose, \\
      & && \transposeInverse,\symetricTranspose
\end{align*}
\cbox{
\textit{\textbf{Differential}}
\begin{align*}
\dd_\mvec\dd_{\vvec\Vmat} \KLD = \braT{\dd\mvec}-\pare{\bra{(\mvec-\wvec)^T\Vmatinv}\otimes \Vmatinv}\dd\vvec{\Vmat}
\end{align*}
}
From where:
\cbox{
\textit{\textbf{Hessian}}
\begin{align*}
    \frac{\dd \KLD}{\dd\mvec^T \dd \vvec\Vmat} = -\bra{(\mvec-\wvec)^T\Vmatinv}\otimes \Vmatinv \in \realNNsquare
\end{align*}
}
Note that this is not explicitly the Hessian from any function, thus, we cannot expect symmetry to apply, something easily verified by inspecting the dimensions from the partial derivative.

Considering only the unique elements from $\Vmat$ we have:
\begin{align*}
\dd_\mvec\dd_{\Vmat} \KLD &= \braT{\dd\mvec}-\pare{\bra{(\mvec-\wvec)^T\Vmatinv}\otimes \Vmatinv}\dd\vvec{\Vmat} \\
&= \braT{\dd\mvec}-\pare{\bra{(\mvec-\wvec)^T\Vmatinv}\otimes \Vmatinv}\Dn\dd\vvech{\Vmat} && \text{using \eqns} \ref{vec2vech}, \ref{dAX}
\end{align*}
\cbox{
\textit{\textbf{Differential}}
\begin{align*}
\dd_\mvec\dd_{\vvech\Vmat} \KLD = \braT{\dd\mvec}-\pare{\bra{(\mvec-\wvec)^T\Vmatinv}\otimes \Vmatinv}\Dn\dd\vvech{\Vmat} 
\end{align*}
}
where:
\cbox{
\textit{\textbf{Hessian}}
\begin{align*}
    \frac{\dd \KLD}{\dd\mvec^T \dd \vvech\Vmat} = -\bra{(\mvec-\wvec)^T\Vmatinv}\otimes \Vmatinv \Dn \in \realNNDup
\end{align*}
}
\subsection{\texorpdfstring{Partial derivative: $\frac{\dd \KLD}{\dd \wvec^T\,\dd\mvec}$}{Partial dw dm}}
Here we deal with $\phi:\realN \times \realN \to \real$.

The first differential \wrt $\mvec$ is:
\begin{align*}
    \dd_\mvec \KLD = (\mvec-\wvec)^T\Vmatinv \dd\mvec
\end{align*}
The second differential \wrt $\wvec$ is:
\begin{align*}
    \dd_\wvec\dd_\mvec \KLD &= \dd_\wvec\bra{(\mvec-\wvec)^T\Vmatinv \dd\mvec}&& \\
    &= \dd_\wvec\bra{\cancelto{0}{\mvec^T\Vmatinv\dd\mvec}}- \dd_\wvec\bra{\wvec^T\Vmatinv \dd\mvec}&& \text{using \eqns } \ref{dconst},\ref{dsum}\\
    &= \braT{\dd\wvec}-\Vmatinv \dd\mvec&& \text{using \eqns } \ref{dXT},\ref{dAX}
\end{align*}
\cbox{
\textit{\textbf{Differential}}
\begin{align*}
    \dd_\wvec\dd_\mvec \KLD = \braT{\dd\wvec}-\Vmatinv \dd\mvec
\end{align*}
}

So we identify:
\cbox{
\textit{\textbf{Hessian}}
\begin{align*}
    \frac{\dd \KLD}{\dd \wvec^T\,\dd\mvec} = -\Vmatinv \in \realNN
\end{align*}
}
which is a symmetric matrix, yet not identified as the Hessian from any function.
\subsection{\texorpdfstring{Partial derivative $\frac{\dd \KLD}{\dd \wvec^T\,\dd\wvec}$}{partial dw dw}}
Here we deal with $\phi: \realN \times \realN \to \real$. 
The Hessian \wrt $\wvec$ is obtained in a similar way to that of $\mvec$. Since the first differential is:
\begin{align*}
    \dd_\wvec \KLD = -(\mvec-\wvec)^T\Vmatinv \Imat\dd\wvec
\end{align*}
The second differential is given by:
\begin{align*}
    \dd^2_\wvec \KLD &= \ddB{-(\mvec-\wvec)^T\Vmatinv\dd\wvec} &&\\
    &=\ddB{-(\mvec-\wvec)^T}\Vmatinv \dd\wvec - (\mvec-\wvec)^T\Vmatinv \cancelto{0}{\ddB{\dd\wvec}} &&\text{using \eqns } \ref{dprod},\ref{dAX},\ref{d2X}\\
    &= \Tdd{-(\mvec-\wvec)}\Vmatinv \dd\wvec && \text{using \eqn } \ref{dXT}\\
    &= \Tdd{\wvec}\Vmatinv \dd\wvec && \text{using \eqns } \ref{dsum},\ref{dconst}
\end{align*}
\cbox{
\textit{\textbf{Differential}}
\begin{align*}
    \dd^2_\wvec \KLD = \Tdd{\wvec}\Vmatinv \dd\wvec 
\end{align*}
}
from where:
\cbox{
\textit{\textbf{Hessian}}
\begin{align*}
   \frac{\dd \KLD}{\dd \wvec^T\,\dd\wvec} = \Vmatinv \in \realNN
\end{align*}
}
which is symmetric. This would be the Hessian of a function considering only $\wvec$ as the argument being differentiated.
\subsection{\texorpdfstring{Partial derivative: $\frac{\dd \KLD}{\dd\wvec^T\,\vvec{\Smat}}$}{Partial dw dS}}
Here we deal with $\phi: \realN \times \mathbb{S}^N_{++} \to \real$.

Because the first differential is:
\begin{align*}
    \dd_\Smat\KLD = \frac{1}{2}\vvecBT{\Vmatinv-\Smatinv}\dd\vvec{\Smat}
\end{align*}
The second differential \wrt $\wvec$ is:
\begin{align*}
    \dd_\wvec\dd_\Smat \KLD &= \dd_\wvec\bra{\frac{1}{2}\vvecBT{\Vmatinv-\Smatinv}\dd\vvec{\Smat}}\\
    &= \zeromat  && \text{since none of the terms depends on } \wvec
\end{align*}
\cbox{
\textit{\textbf{Differential}}
\begin{align*}
\dd_\wvec\dd_{\vvec\Smat} \KLD = \braT{\dd \wvec} \zeromat \dd\vvec{\Smat} 
\end{align*}
}
So we identify:
\cbox{
\textit{\textbf{Hessian}}
\begin{align*}
    \frac{\dd \KLD}{\dd\wvec^T\,\dd\vvec{\Smat}} = \zeromat \in \realNNsquare
\end{align*}
}
The resulting partial derivative does not represent the Hessian of any function, which is why it is not a symmetric matrix.

Considering only the unique elements of $\Smat$ we have:

\cbox{
\textit{\textbf{Differential}}
\begin{align*}
    \dd_\wvec \dd_{\vvech{\Smat}} \KLD &= \braT{\dd \wvec} \zeromat \Dn \dd\vvech{\Smat} && \text{using \eqns } \ref{vec2vech}, \ref{dAX}
\end{align*}
}
\cbox{
\textit{\textbf{Hessian}}
\begin{align*}
    \frac{\dd \KLD}{\dd\wvec^T\,\dd\vvech{\Smat}} = \zeromat \Dn \in \realNNDup
\end{align*}
}
\subsection{\texorpdfstring{Partial derivative $\frac{\dd \KLD}{\dd\wvec^T \dd \vvec\Vmat}$}{Partial dw dV}}
Here we deal with $\phi: \realN \times \mathbb{S}^N_{++} \to \real$.

Because:
\begin{align*}
      \dd_\Vmat \KLD &=\frac{1}{2}\bra{\trB{\Vmatinv\dd\Vmat}-\trB{\Vmatinv\Smat\Vmatinv\dd\Vmat}-\trB{\Vmatinv\avec\avec^T\Vmatinv\dd\Vmat}} ; \avec = (\mvec - \wvec)
\end{align*}
then:
\begin{align*}
      \dd_\wvec\dd_\Vmat \KLD &= \dd_\wvec\bra{\frac{1}{2}\pare{\cancelto{0}{\trB{\Vmatinv\dd\Vmat}}-\cancelto{0}{\trB{\Vmatinv\Smat\Vmatinv\dd\Vmat}}-\trB{\Vmatinv\avec\avec^T\Vmatinv\dd\Vmat}}} \\
      &=-\frac{1}{2}\dd_\wvec\pare{\trB{\avec^T\Vmatinv\dd\Vmat\Vmatinv\avec}}&& \text{using \eqns } \ref{dsum},\ref{dconst},\circularTR\\
      &=-\frac{1}{2}\trB{\ddB{\avec^T}\Vmatinv\dd\Vmat\Vmatinv\avec + \avec^T\Vmatinv\dd\Vmat\Vmatinv\dd\avec}&& \text{using \eqns } \ref{dprod},\ref{dAX},\ref{dTR}\\
      &=-\frac{1}{2}\trB{\avec^T\Vmatinv\dd\Vmat\Vmatinv\dd\avec + \avec^T\Vmatinv\dd\Vmat\Vmatinv\dd\avec}&& \text{using \eqn } \ref{dXT},\quadraticScalar,\scalarTranspose,\\
      & && \prodTranspose,\transposeInverse,\\
      & && \symetricTranspose\\
      &=\trB{(\mvec-\wvec)^T\Vmatinv\dd\Vmat\Vmatinv\dd\wvec}&& \text{using \eqns } \ref{dsum},\ref{dconst}\\
     &=\trB{\braT{\dd\wvec}\Vmatinv\dd\Vmat\Vmatinv(\mvec-\wvec)}&& \text{using } ,\quadraticScalar,\scalarTranspose,\\
      & && \prodTranspose,\transposeInverse,\\
      & && \symetricTranspose\\
      &=\trB{\Vmatinv(\mvec-\wvec)\braT{\dd\wvec}\Vmatinv\dd\Vmat}&& \text{using }\circularTR \\
      &=\vvecBT{\dd\wvec}\pare{\braT{\Vmatinv(\mvec-\wvec)}\otimes \Vmatinv}\vvec{\dd\Vmat}&& \text{using \eqn} \ref{tr_ABCD_2vec} \\
      &=\braT{\dd\wvec}\pare{\bra{(\mvec-\wvec)^T\Vmatinv}\otimes \Vmatinv}\dd\vvec{\Vmat}&& \text{using \eqns } \ref{vec_of_a_vec},\ref{dvec},\prodTranspose\\
      & && \transposeInverse,\symetricTranspose
\end{align*}
\cbox{
\textit{\textbf{Differential}}
\begin{align*}
    \dd_\wvec\dd_{\vvec\Vmat} \KLD &= \braT{\dd\wvec}\pare{\bra{(\mvec-\wvec)^T\Vmatinv}\otimes \Vmatinv}\dd\vvec{\Vmat}
\end{align*}
}
From where:
\cbox{
\textit{\textbf{Hessian}}
\begin{align*}
    \frac{\dd \KLD}{\dd\wvec^T \dd \vvec\Vmat} = \bra{(\mvec-\wvec)^T\Vmatinv} \otimes \Vmatinv \in \realNNsquare
\end{align*}
}
which is a non-symmetric matrix since this expression does not represent the Hessian from any possible function.
Considering only the unique elements of $\Vmat$ we have:
\begin{align*}
    \dd_\wvec\dd_\Vmat \KLD &= \braT{\dd\wvec}\pare{\bra{(\mvec-\wvec)^T\Vmatinv}\otimes \Vmatinv}\dd\vvec{\Vmat}\\
    &=  \braT{\dd\wvec}\pare{\bra{(\mvec-\wvec)^T\Vmatinv}\otimes \Vmatinv}\Dn\dd\vvech{\Vmat} && \text{using \eqns } \ref{vec2vech},\ref{dAX}
\end{align*}
\cbox{
\textit{\textbf{Differential}}
\begin{align*}
    \dd_\wvec\dd_{\vvech\Vmat} \KLD &=  \braT{\dd\wvec}\pare{\bra{(\mvec-\wvec)^T\Vmatinv}\otimes \Vmatinv}\Dn\dd\vvech{\Vmat} 
\end{align*}
}
from where:
\cbox{
\textit{\textbf{Hessian}}
\begin{align*}
    \frac{\dd \KLD}{\dd\wvec^T \dd \vvech\Vmat} = \bra{(\mvec-\wvec)^T\Vmatinv} \otimes \Vmatinv \Dn \in \realNNDup
\end{align*}
}
\subsection{\texorpdfstring{Partial derivative: $\frac{\dd \KLD}{\dd \vvecT{\Smat}\,\dd\mvec}$}{Partial dS dm}}
Here we deal with $\phi: \mathbb{S}^N_{++} \times  \realN \to \real$. Because the first differential is given by:
\begin{align*}
    \dd_\mvec \KLD = (\mvec-\wvec)^T\Vmatinv \dd\mvec
\end{align*}
The second differential \wrt $\Smat$ is:
\begin{align*}
    \dd_\Smat\dd_\mvec \KLD &= \dd_\Smat\bra{(\mvec-\wvec)^T\Vmatinv \dd\mvec}\\
    &= \zeromat && \text{since none of the terms depends on } \Smat
\end{align*}
\cbox{
\textit{\textbf{Differential}}
\begin{align*}
    \dd_{\vvec\Smat}\dd_\mvec \KLD &= \braT{\dd\vvec \Smat} \zeromat \dd\mvec 
\end{align*}
}
So we identify:
\cbox{
\textit{\textbf{Hessian}}
\begin{align*}
    \frac{\dd \KLD}{\dd \vvecT{\Smat}\,\dd\mvec} = \zeromat \in \realNsquareN
\end{align*}
}
which is not symmetric as it does not represent the Hessian from any possible function.

Considering only the unique elements of $\Smat$ we have:
\cbox{
\textit{\textbf{Differential}}
\begin{align*}
    \dd_{\vvech\Smat}\dd_\mvec \KLD &= \braT{\dd\vvech \Smat} \DnT \zeromat \dd\mvec && \text{using \eqns } \ref{vec2vech},\ref{dAX},\prodTranspose
\end{align*}
}
\cbox{
\textit{\textbf{Hessian}}
\begin{align*}
    \frac{\dd \KLD}{\dd \vvechT{\Smat}\,\dd\mvec} =\DnT \zeromat \in \realNDupN
\end{align*}
}
\subsection{\texorpdfstring{Partial derivative: $\frac{\dd \KLD}{\dd \vvecT{\Smat}\,\dd\wvec}$}{Partial dS dw}}
Here we deal with $\phi: \mathbb{S}^N_{++} \times  \realN \to \real$. Because the first differential is given by:
\begin{align*}
     \dd_\wvec \KLD = -(\mvec-\wvec)^T\Vmatinv \dd\wvec
\end{align*}
The second differential \wrt $\Smat$ is:
\begin{align*}
    \dd_\Smat\dd_\wvec \KLD &= \dd_\Smat\bra{-(\mvec-\wvec)^T\Vmatinv \dd\wvec}\\
    &= \zeromat  && \text{since none of the terms depends on } \Smat
\end{align*}
\cbox{
\textit{\textbf{Differential}}
\begin{align*}
    \dd_{\vvec\Smat}\dd_\wvec \KLD &= \braT{\dd\vvec \Smat} \zeromat \dd\wvec 
\end{align*}
}
So we identify:
\cbox{
\textit{\textbf{Hessian}}
\begin{align*}
    \frac{\dd \KLD}{\dd \vvecT{\Smat}\,\dd\wvec} = \zeromat \in \realNsquareN
\end{align*}
}
The resulting partial derivative does not represent the Hessian of any function, which is why it is not a symmetric matrix.

Considering only the unique elements of $\Smat$ we have:
\cbox{
\textit{\textbf{Differential}}
\begin{align*}
    \dd_{\vvech\Smat}\dd_\wvec \KLD &= \braT{\dd\vvech \Smat} \DnT \zeromat \dd\wvec && \text{using \eqns } \ref{vec2vech},\ref{dAX},\prodTranspose
\end{align*}
}
\cbox{
\textit{\textbf{Hessian}}
\begin{align*}
    \frac{\dd \KLD}{\dd \vvechT{\Smat}\,\dd\wvec} =\DnT \zeromat \in \realNDupN
\end{align*}
}
\subsection{\texorpdfstring{Partial derivative: $\frac{\dd \KLD}{\dd \vvecT{\Smat}\,\dd\vvec\Smat}$}{partial dS dS}}
Here we deal with $\phi : \mathbb{S}^N_{++} \times  \mathbb{S}^N_{++} \to \real$.
\subsubsection{From the second-order differential}
The first-order differential is:
\begin{align*}
    \dd_\Smat\KLD = \frac{1}{2}\vvecBT{\Vmatinv-\Smatinv}\dd\vvec{\Smat}
\end{align*}
So the second-order differential is given by computing the differential of the first-order differential:
\begin{align*}
    \dd^2_\Smat\KLD &= \ddB{\frac{1}{2}\vvecBT{\Vmatinv-\Smatinv}\dd\vvec{\Smat}}  &&\\
    &= \frac{1}{2}\ddB{\vvecBT{\Vmatinv-\Smatinv}}\dd\vvec{\Smat} + \frac{1}{2}\vvecBT{\Vmatinv-\Smatinv}\cancelto{0}{\ddB{\dd\vvec{\Smat}}} && \text{using \eqns } \ref{dprod},\ref{d2X}\\
    &= \frac{1}{2}\braT{\ddB{\vvecB{\Vmatinv-\Smatinv}}}\dd\vvec{\Smat} && \text{using \eqn } \ref{dXT}\\
    &= -\frac{1}{2}\braT{\vvecB{\dd\Smatinv}}\dd\vvec{\Smat} && \text{using \eqns } \ref{dsum},\ref{dconst},\ref{dvec}\\
    &= \frac{1}{2}\pareT{\vvecB{\Smatinv\dd\Smat\Smatinv}}\dd\vvec{\Smat} && \text{using \eqn } \ref{dinv}\\
    &= \frac{1}{2}\pareT{\bra{\Smatinvt\otimes \Smatinv} \vvec{\dd\Smat}}\dd\vvec{\Smat} && \text{using \eqn } \ref{vecmat2kron}\\
    &= \frac{1}{2}\braT{\dd\vvec{\Smat}}\braT{\Smatinv\otimes \Smatinv} \dd\vvec{\Smat} && \text{using \eqn } \ref{dvec}, \prodTranspose,\\
    & && \transposeInverse,\symetricTranspose\\
    &= \braT{\dd\vvec{\Smat}}\frac{1}{2}\bra{\Smatinv\otimes \Smatinv} \dd\vvec{\Smat} && \text{using } \kronTranspose,\\
    & && \transposeInverse,\symetricTranspose
\end{align*}
\cbox{
\textit{\textbf{Differential }}
\begin{align*}
\dd^2_{\vvec\Smat} \KLD  = \braT{\dd\vvec{\Smat}}\frac{1}{2}\bra{\Smatinv\otimes \Smatinv} \dd\vvec{\Smat}
\end{align*}
}
So we identify:
\cbox{
\textit{\textbf{Hessian}}
\begin{align*}
    \frac{\dd \KLD}{\dd \vvecT{\Smat}\,\dd\vvec\Smat} = \frac{1}{2}\bra{\Smatinv\otimes \Smatinv} \in \realNsquareNsquare
\end{align*}
}
This is a symmetric matrix, as expected, since it is the Hessian from the \KLD function considering only $\Smat$ as a variable.

Considering only the unique elements from $\Smat$ we have:
\begin{align*}
\dd^2_{\Smat} \KLD  &= \braT{\dd\vvec{\Smat}}\frac{1}{2}\bra{\Smatinv\otimes \Smatinv} \dd\vvec{\Smat} &&\\
&=  \braT{\Dn\dd\vvech{\Smat}}\frac{1}{2}\bra{\Smatinv\otimes \Smatinv} \Dn\dd\vvech{\Smat} && \text{using \eqn} \ref{vec2vech},\ref{dAX}\\
&=  \braT{\dd\vvech{\Smat}}\DnT\frac{1}{2}\bra{\Smatinv\otimes \Smatinv} \Dn\dd\vvech{\Smat} && \text{using } \prodTranspose
\end{align*}
\cbox{
\textit{\textbf{Differential}}
\begin{align*}
\dd^2_{\vvech\Smat} \KLD  = \braT{\dd\vvech{\Smat}}\DnT\frac{1}{2}\bra{\Smatinv\otimes \Smatinv} \Dn\dd\vvech{\Smat}
\end{align*}
}
from where:
\cbox{
\textit{\textbf{Hessian}}
\begin{align*}
\frac{\dd \KLD}{\dd \vvechT{\Smat}\,\dd\vvech\Smat}  = \DnT\frac{1}{2}\bra{\Smatinv\otimes \Smatinv} \Dn\in \realNDupNDup
\end{align*}
}
\subsubsection{From the second order differential through a different expression}
Note that the first-order differential can be written as:
\begin{align*}
    \dd_\Smat\KLD &= \frac{1}{2}\vvecBT{\Vmatinv-\Smatinv}\dd\vvec{\Smat}\\
    &= \frac{1}{2}\vvecBT{\Vmatinv-\Smatinv}\vvec{\dd\Smat} && \text{using \eqn } \ref{dvec} \\
    &= \frac{1}{2}\trB{\pareT{\Vmatinv-\Smatinv}\dd\Smat}&& \text{using \eqn } \ref{tr_matprod_2vec}\\
    &= \frac{1}{2}\trB{\pare{\Vmatinv-\Smatinv}\dd\Smat}&& \text{using } \sumTranspose, \transposeInverse,\symetricTranspose.
\end{align*}
We now obtain the second differential of this expression:
\begin{align*}
    \dd^2_\Smat\KLD &=  \ddB{\frac{1}{2}\trB{\pare{\Vmatinv-\Smatinv}\dd\Smat}}\\
    &=  \frac{1}{2}\trB{\ddB{\pare{\Vmatinv-\Smatinv}\dd\Smat}} && \text{using \eqn } \ref{dTR}\\
    &=  \frac{1}{2}\trB{\ddB{\pare{\Vmatinv-\Smatinv}}\dd\Smat + \pare{\Vmatinv-\Smatinv}\cancelto{0}{\ddB{\dd\Smat}}}&& \text{using \eqns } \ref{dprod},\ref{d2X}\\
    &=  \frac{1}{2}\trB{-\dd\Smatinv\dd\Smat} && \text{using \eqns } \ref{dsum},\ref{dconst}\\
    &=  \trB{\frac{1}{2}\Smatinv\dd\Smat\Smatinv\dd\Smat} && \text{using \eqn } \ref{dinv}\\
    &=  \trB{\frac{1}{2}\Smatinv\braT{\dd\Smat}\Smatinv\dd\Smat} && \text{using \eqn } \ref{dXT}, \symetricTranspose
\end{align*}
Identifying the Hessian as (since remember this submatrix is directly the Hessian from a function considering only $\Smat$ as a function argument):
\begin{align*}
    \frac{\dd \KLD}{\dd \vvecT{\Smat}\dd \vvec{\Smat}} &= \frac{1}{2}\pare{\frac{1}{2}\Smatinvt \otimes \Smatinv + \frac{1}{2}\Smatinv \otimes \Smatinvt}&& \text{using \eqn } \ref{tr_dXdX_2Hess}\\
    &= \frac{1}{2}\pare{\frac{1}{2}\pare{\Smatinv \otimes \Smatinv} + \frac{1}{2}\pare{\Smatinv \otimes \Smatinv}}&& \text{using } \transposeInverse,\symetricTranspose,\kronScalar\\
    &= \frac{1}{2}\pare{\Smatinv \otimes \Smatinv} \in \realNsquareNsquare&& 
\end{align*}
as expected. However, rather than operating through the theorem, we could continue operating on the expression till yield the canonical form. So:
\begin{align*}
    \dd^2_\Smat\KLD &=  \hdots\\
    &=  \trB{\frac{1}{2}\Smatinv\dd\Smat\Smatinv\dd\Smat} && \\
    &=  \frac{1}{2}\trB{\dd\Smat\Smatinv\dd\Smat\Smatinv} && \text{using }\circularTR\\
    &=  \frac{1}{2}\vvecBT{\braT{\dd{\Smat}}}\vvecB{\Smatinv\dd\Smat\Smatinv} && \text{using \eqn } \ref{tr_matprod_2vec}\\
    &=  \frac{1}{2}\braT{\dd{\vvec{\Smat}}}\vvecB{\Smatinv\dd\Smat\Smatinv} && \text{using \eqn } \ref{dXT},\symetricTranspose,\text{\eqn}\, \ref{dvec}\\
    &=  \braT{\dd{\vvec{\Smat}}}\frac{1}{2}\bra{\Smatinv\otimes\Smatinv}\dd\vvec{\Smat}&& \text{using \eqns } \ref{vecmat2kron},\ref{dvec},\transposeInverse,\symetricTranspose
\end{align*}
From where we can identify the Hessian:
\begin{align*}
    \frac{\dd \KLD}{\dd \vvecT{\Smat}\dd \vvec{\Smat}} = \frac{1}{2}\bra{\Smatinv\otimes \Smatinv} \in \real^{N^2 \times N^2}
\end{align*}
as expected.
\subsubsection{From the first order differential of the transposed Jacobian}
The Jacobian is:
\begin{align*}
    \Jacobian{\KLD,\vvec{\Smat}} = \frac{1}{2}\vvecBT{\Vmatinv-\Smatinv}  \in \realNsquarerow
\end{align*}
Now because:
\begin{align*}
    \frac{\dd \KLD}{\dd \vvecT{\Smat}\dd \vvec{\Smat}} = \frac{\dd}{\braT{\dd \vvec{\Smat}}} \braT{\Jacobian{\KLD,\vvec{\Smat}}}
\end{align*}
Thus from the first-order differential of the transposed Jacobian:
\begin{align*}
   \dd_\Smat \Jacobian{\KLD,\vvec{\Smat}}^T &= \dd_\Smat\braT{\frac{1}{2}\vvecBT{\Vmatinv-\Smatinv}}&&\\
&=\frac{1}{2}\vvecB{\cancelto{0}{\dd_\Smat\Vmatinv}-\dd\Smatinv}&& \text{using \eqns } \ref{dvec},\ref{dconst},\ref{dsum}\\
&= \frac{1}{2}\vvecB{\Smatinv\dd\Smat\Smatinv}&& \text{using \eqn } \ref{dinv}\\
&= \frac{1}{2}\bra{\Smatinv\otimes\Smatinv}\dd\vvec{\Smat}&& \text{using \eqns } \ref{vecmat2kron},\ref{dvec},\transposeInverse,\symetricTranspose
\end{align*}
from which the Hessian is identified as being the Jacobian of this differential. Thus:
\begin{align*}
    \frac{\dd \KLD}{\dd \vvecT{\Smat}\dd \vvec{\Smat}} = \frac{1}{2}\bra{\Smatinv\otimes \Smatinv} \in \real^{N^2 \times N^2}
\end{align*}
as expected.
\subsection{\texorpdfstring{Partial derivative $\frac{\dd \KLD}{\dd\vvecT{\Smat},\dd\vvec{\Vmat}}$}{Partial dS dV}}
Here we deal with $\phi: \mathbb{S}^N_{++} \times \mathbb{S}^N_{++} \to \real$. The first differential \wrt $\Vmat$ is given by:
\begin{align*}
      \dd_\Vmat \KLD &=\frac{1}{2}\pare{\trB{\Vmatinv\dd\Vmat}-\trB{\Vmatinv\Smat\Vmatinv\dd\Vmat}-\trB{\Vmatinv\avec\avec^T\Vmatinv\dd\Vmat}}
\end{align*}
The second differential \wrt $\Smat$ is given by:
\begin{align*}
      \dd_\Smat\dd_\Vmat \KLD &=\dd_\Smat\bra{\frac{1}{2}\pare{\cancelto{0}{\trB{\Vmatinv\dd\Vmat}}-\trB{\Vmatinv\Smat\Vmatinv\dd\Vmat}-\cancelto{0}{\trB{\Vmatinv\avec\avec^T\Vmatinv\dd\Vmat}}}} && \\
      &=\trB{-\frac{1}{2}\Vmatinv\dd\Smat\Vmatinv\dd\Vmat} && \text{using \eqns } \ref{dconst},\ref{dTR},\ref{dsum},\ref{dAX}\\
      &=\trB{-\frac{1}{2}\Vmatinv\braT{\dd\Smat}\Vmatinv\dd\Vmat} &&\text{using } \symetricTranspose,\text{\eqn }\ref{dXT}
\end{align*}
Invoking theorem yields:
\begin{align*}
    \frac{\dd \KLD}{\dd\vvecT{\Smat},\dd\vvec{\Vmat}} &= \frac{1}{2}\pare{\braT{-\frac{1}{2}\Vmatinv}\otimes \Vmatinv -\frac{1}{2}\Vmatinv\otimes\Vmatinvt }&& \text{using \eqn } \ref{tr_dYdX_2Hess}\\
    &=\frac{1}{2}\pare{-\frac{1}{2}\Vmatinv\otimes \Vmatinv  -\frac{1}{2}\Vmatinv\otimes\Vmatinv }&& \text{using } \symetricTranspose,\transposeInverse\\
    &=-\frac{1}{2}\pare{\Vmatinv\otimes \Vmatinv}&& \text{using }\kronScalar 
\end{align*}
\cbox{
\textit{\textbf{Differential}}
\begin{align*}
    \dd_{\vvec\Smat}\dd_{\vvec\Vmat} \KLD &=\braT{\dd\vvec{}\Smat} -\frac{1}{2}\pare{\Vmatinv\otimes \Vmatinv}\dd\vvec{}\Vmat
\end{align*}
}
from where:
\cbox{
\textit{\textbf{Hessian}}
\begin{align*}
  \frac{\dd \KLD}{\dd\vvecT{\Smat},\dd\vvec{\Vmat}} &= 
    -\frac{1}{2}\pare{\Vmatinv\otimes \Vmatinv} \in \realNsquareNsquare
\end{align*}
}
While this expression is a symmetric matrix, it does not represent the Hessian from any function.
Considering only unique elements from both $\Smat,\Vmat$, then:
\begin{align*}
    \dd_{\Smat}\dd_{\Vmat} \KLD &=\braT{\dd\vvec{}\Smat} -\frac{1}{2}\pare{\Vmatinv\otimes \Vmatinv}\dd\vvec{}\Vmat \\
    &= \braT{\Dn\dd\vvech{}\Smat} -\frac{1}{2}\pare{\Vmatinv\otimes \Vmatinv}\Dn\dd\vvech{}\Vmat && \text{using \eqns} \ref{vec2vech},\ref{dAX}\\
    &= \braT{\dd\vvech{}\Smat}\DnT -\frac{1}{2}\pare{\Vmatinv\otimes \Vmatinv}\Dn\dd\vvech{}\Vmat && \text{using } \prodTranspose
\end{align*}
\cbox{
\textit{\textbf{Differential}}
\begin{align*}
    \dd_{\vvech\Smat}\dd_{\vvech\Vmat} \KLD &=\braT{\dd\vvech{}\Smat} \DnT-\frac{1}{2}\pare{\Vmatinv\otimes \Vmatinv}\Dn\dd\vvech{}\Vmat
\end{align*}
}
from where:
\cbox{
\textit{\textbf{Hessian}}
\begin{align*}
  \frac{\dd \KLD}{\dd\vvechT{\Smat},\dd\vvech{\Vmat}} &= 
    \DnT-\frac{1}{2}\pare{\Vmatinv\otimes \Vmatinv}\Dn \in \realNDupNDup
\end{align*}
}
\subsection{\texorpdfstring{Partial derivative $\frac{\dd \KLD}{\dd \vvecT{\Vmat}\dd\mvec}$}{Partial dV dm}}
Here we deal with $\phi : \mathbb{S}^N_{++} \times \realN \to \real$.

Because:
\begin{align*}
    \dd_\mvec \KLD = (\mvec-\wvec)^T\Vmatinv \dd\mvec
\end{align*}
we have:
\begin{align*}
    \dd_\Vmat \dd_\mvec \KLD &= \dd_\Vmat\bra{(\mvec-\wvec)^T\Vmatinv \dd\mvec}\\
    &= (\mvec-\wvec)^T\dd\Vmatinv \dd\mvec&& \text{using \eqn } \ref{dAX}\\
    &= -(\mvec-\wvec)^T\Vmatinv\dd\Vmat\Vmatinv \dd\mvec&& \text{using \eqn } \ref{dinv}\\
    &= -\trB{(\mvec-\wvec)^T\Vmatinv\dd\Vmat\Vmatinv \dd\mvec}&& \text{using }\quadraticScalar,\scalarTR \\
    &= -\trB{\dd\Vmat\Vmatinv \dd\mvec(\mvec-\wvec)^T\Vmatinv}&& \text{using }\circularTR \\
    &= \vvecT{\braT{\dd\Vmat}}-\braT{(\mvec-\wvec)^T\Vmatinv}\otimes\Vmatinv \dd\mvec&& \text{using \eqn} \ref{tr_ABcdT_2vec}\\
    &= \braT{\dd\vvec{\Vmat}}-\bra{\Vmatinv(\mvec-\wvec)}\otimes\Vmatinv \dd\mvec&& \text{using \eqns } \ref{dvec},\ref{dXT},\prodTranspose,\\
    & && \symetricTranspose,\transposeInverse
\end{align*}
\cbox{
\textit{\textbf{Differential}}
\begin{align*}
     \dd_{\vvec\Vmat} \dd_\mvec \KLD = \braT{\dd\vvec{\Vmat}}-\bra{\Vmatinv(\mvec-\wvec)}\otimes\Vmatinv \dd\mvec
\end{align*}
}
From where we identify:
\cbox{
\textit{\textbf{Hessian}}
\begin{align*}
    \frac{\dd \KLD}{\dd \vvecT{\Vmat}\dd\mvec} &= -\bra{\Vmatinv(\mvec-\wvec)}\otimes\Vmatinv \in \realNsquareN
\end{align*}
}
which is a non-symmetric matrix since it does not represent the Hessian from any function.

Considering only the unique elements from $\Vmat$ we have:
\begin{align*}
     \dd_{\Vmat} \dd_\mvec \KLD = \braT{\dd\vvec{\Vmat}}-\bra{\Vmatinv(\mvec-\wvec)}\otimes\Vmatinv \dd\mvec\\
     \braT{\Dn\dd\vvech{\Vmat}}-\bra{\Vmatinv(\mvec-\wvec)}\otimes\Vmatinv \dd\mvec && \text{using \eqns } \ref{vec2vech},\ref{dAX} \\
     \braT{\dd\vvech{\Vmat}}\DnT-\bra{\Vmatinv(\mvec-\wvec)}\otimes\Vmatinv \dd\mvec && \text{using } \prodTranspose
\end{align*}
\cbox{
\textit{\textbf{Differential}}
\begin{align*}
     \dd_{\vvech\Vmat} \dd_\mvec \KLD = \braT{\dd\vvech{\Vmat}}\DnT-\bra{\Vmatinv(\mvec-\wvec)}\otimes\Vmatinv \dd\mvec
\end{align*}
}
from where:
\cbox{
\textit{\textbf{Hessian}}
\begin{align*}
    \frac{\dd \KLD}{\dd \vvechT{\Vmat}\dd\mvec} &= \DnT-\bra{\Vmatinv(\mvec-\wvec)}\otimes\Vmatinv \in \realNDupN
\end{align*}
}
\subsection{\texorpdfstring{Partial derivative $\frac{\dd \KLD}{\dd \vvecT{\Vmat}\dd\wvec}$}{Partial dV dw}}
Here we deal with $\phi : \mathbb{S}^N_{++} \times \realN  \to \real$. Because:
\begin{align*}
     \dd_\wvec \KLD = -(\mvec-\wvec)^T\Vmatinv \dd\wvec
\end{align*}
we have:
\begin{align*}
    \dd_\Vmat \dd_\wvec \KLD &= \dd_\Vmat\bra{-(\mvec-\wvec)^T\Vmatinv \dd\wvec}\\
    &= -(\mvec-\wvec)^T\dd\Vmatinv \dd\wvec&& \text{using \eqn } \ref{dAX}\\
    &= (\mvec-\wvec)^T\Vmatinv\dd\Vmat\Vmatinv \dd\wvec&& \text{using \eqn } \ref{dinv}\\
    &= \trB{(\mvec-\wvec)^T\Vmatinv\dd\Vmat\Vmatinv \dd\wvec}&& \text{using } \quadraticScalar,\scalarTR\\
    &= \trB{\dd\Vmat\Vmatinv \dd\wvec(\mvec-\wvec)^T\Vmatinv}&& \text{using } \circularTR\\
    &= \vvecT{\braT{\dd\Vmat}}\braT{(\mvec-\wvec)^T\Vmatinv}\otimes\Vmatinv {\dd\wvec}&& \text{using \eqn } \ref{tr_ABcdT_2vec}\\
    &= \braT{\dd\vvec{\Vmat}}\bra{\Vmatinv(\mvec-\wvec)}\otimes\Vmatinv \dd\wvec&& \text{using \eqns } \ref{dvec},\ref{dXT},\prodTranspose,\\
    & && \symetricTranspose,\transposeInverse
\end{align*}
\cbox{
\textit{\textbf{Differential}}
\begin{align*}
    \dd_{\vvec\Vmat} \dd_\wvec \KLD &=  \braT{\dd\vvec{\Vmat}}\bra{\Vmatinv(\mvec-\wvec)}\otimes\Vmatinv \dd\wvec
\end{align*}
}
From where we identify:
\cbox{
\textit{\textbf{Hessian}}
\begin{align*}
    \frac{\dd \KLD}{\dd \vvecT\Vmat\dd\wvec} &= \bra{\Vmatinv(\mvec-\wvec)}\otimes\Vmatinv \in \realNsquareN
\end{align*}
}
which is a non-symmetric matrix since it does not represent the Hessian from any function.

Considering only the unique elements from $\Vmat$ we have:
\begin{align*}
    \dd_\Vmat \dd_\wvec \KLD &=  \braT{\dd\vvec{\Vmat}}\bra{\Vmatinv(\mvec-\wvec)}\otimes\Vmatinv \dd\wvec \\
    &= \braT{\Dn\dd\vvech{\Vmat}}\bra{\Vmatinv(\mvec-\wvec)}\otimes\Vmatinv \dd\wvec && \text{using \eqns }\ref{vec2vech},\ref{dAX}\\
    &= \braT{\dd\vvech{\Vmat}}\DnT\bra{\Vmatinv(\mvec-\wvec)}\otimes\Vmatinv \dd\wvec && \text{using } \prodTranspose
\end{align*}
\cbox{
\textit{\textbf{Differential}}
\begin{align*}
    \dd_{\vvech\Vmat} \dd_\wvec \KLD &=  \braT{\dd\vvech{\Vmat}}\DnT\bra{\Vmatinv(\mvec-\wvec)}\otimes\Vmatinv \dd\wvec
\end{align*}
}
From where we identify:
\cbox{
\textit{\textbf{Hessian}}
\begin{align*}
    \frac{\dd \KLD}{\dd \vvechT\Vmat\dd\wvec} &= \DnT\bra{\Vmatinv(\mvec-\wvec)}\otimes\Vmatinv \in \realNDupN
\end{align*}
}
\subsection{\texorpdfstring{Partial derivative $\frac{\dd \KLD}{\dd\vvecT{\Vmat},\dd\vvec{\Smat}}$}{Partial dV dS}}
Here we deal with $\phi : \mathbb{S}^N_{++} \times \mathbb{S}^N_{++}  \to \real$. The first differential \wrt $\Smat$ is given by:
\begin{align*}
    \dd_\Smat\KLD &= \frac{1}{2}\vvecBT{\Vmatinv-\Smatinv}\dd\vvec{\Smat}\\
    &= \frac{1}{2}\trB{\pareT{\Vmatinv-\Smatinv}\dd\Smat}&& \text{using \eqns } \ref{tr_matprod_2vec},\ref{dvec}\\
    &= \frac{1}{2}\trB{\pare{\Vmatinv-\Smatinv}\dd\Smat}&& \text{using } \sumTranspose,\transposeInverse,\symetricTranspose
\end{align*}
Now the second differential \wrt $\Vmat$:
\begin{align*}
    \dd_\Vmat\dd_\Smat\KLD &= \dd_\Vmat\pare{\frac{1}{2}\trB{\pare{\Vmatinv-\Smatinv}\dd\Smat}}\\
    &=\frac{1}{2}\dd_\Vmat\pare{\trB{\Vmatinv\dd\Smat-\Smatinv\dd\Smat}}&& \\
    &=\frac{1}{2}\trB{\dd_\Vmat\Vmatinv\dd\Smat-\cancelto{0}{\dd_\Vmat\Smatinv\dd\Smat}}&& \text{using \eqns } \ref{dconst},\ref{dTR},\ref{dsum}\\
    &=-\frac{1}{2}\trB{\Vmatinv\dd\Vmat\Vmatinv\dd\Smat}&& \text{using \eqns } \ref{dinv},\ref{dAX}\\
    &=\trB{-\frac{1}{2}\Vmatinv\braT {\dd\Vmat}\Vmatinv\dd\Smat}&& \text{using } \symetricTranspose,\text{\eqn }\ref{dXT}
\end{align*}
So we can apply the theorem and result in:
\begin{align*}
    \frac{\dd \KLD}{\dd\vvecT{\Vmat},\dd\vvec{\Smat}} &= \frac{1}{2}\pare{\braT{-\frac{1}{2}\Vmatinv}\otimes \Vmatinv -\frac{1}{2}\Vmatinv\otimes\Vmatinvt }&& \text{using \eqn } \ref{tr_dYdX_2Hess}\\
    &=\frac{1}{2}\pare{-\frac{1}{2}\Vmatinv\otimes \Vmatinv  -\frac{1}{2}\Vmatinv\otimes\Vmatinv }&& \text{using } \transposeInverse,\symetricTranspose\\
    &=-\frac{1}{2}\pare{\Vmatinv\otimes \Vmatinv }&& \text{using } \kronScalar
\end{align*}
\cbox{
\textit{\textbf{Differential}}
\begin{align*}
    \dd_{\vvec\Vmat}\dd_{\vvec\Smat} = \braT{\dd\vvec\Vmat} -\frac{1}{2}\pare{\Vmatinv\otimes \Vmatinv } \dd\vvec\Smat
\end{align*}
}
\cbox{
\textit{\textbf{Hessian}}
\begin{align*}
    \frac{\dd \KLD}{\dd\vvecT{\Vmat},\dd\vvec{\Smat}} &= -\frac{1}{2}\pare{\Vmatinv\otimes \Vmatinv } \in \realNsquareNsquare
\end{align*}
}
While this expression represents a symmetric matrix, it does not represent the Hessian from any function.

Considering only the unique elements from $\Smat,\Vmat$ we have:
\begin{align*}
    \dd_{\Vmat}\dd_{\Smat} &= \braT{\dd\vvec\Vmat} -\frac{1}{2}\pare{\Vmatinv\otimes \Vmatinv } \dd\vvec\Smat &&\\
    &= \braT{\Dn\dd\vvech\Vmat} -\frac{1}{2}\pare{\Vmatinv\otimes \Vmatinv } \Dn\dd\vvech\Smat && \text{using \eqns } \ref{vec2vech},\ref{dAX}\\
    &= \braT{\dd\vvech\Vmat} \DnT-\frac{1}{2}\pare{\Vmatinv\otimes \Vmatinv } \Dn\dd\vvech\Smat && \text{using } \prodTranspose
\end{align*}
\cbox{
\textit{\textbf{Differential}}
\begin{align*}
\dd_{\vvech\Vmat}\dd_{\vvech\Smat}\KLD = \braT{\dd\vvech\Vmat} \DnT-\frac{1}{2}\pare{\Vmatinv\otimes \Vmatinv } \Dn\dd\vvech\Smat
\end{align*}
}
\cbox{
\textit{\textbf{Hessian}}
\begin{align*}
    \frac{\dd \KLD}{\dd\vvechT{\Vmat},\dd\vvech{\Smat}} &= \DnT-\frac{1}{2}\pare{\Vmatinv\otimes \Vmatinv } \Dn \in \realNDupNDup
\end{align*}
}
\subsection{\texorpdfstring{Partial derivative: $\frac{\dd \KLD}{\dd \vvecT{\Vmat}\,\dd\vvec\Vmat}$}{Partial dV dV}}
Here we deal with $\phi: \mathbb{S}^N_{++} \times \mathbb{S}^N_{++} \to \real$.
\subsubsection{From the second-order differential}
Define $\Cmat = \pare{\Smat+\avec\avec^T},\,\avec=(\mvec-\wvec)$, which is a symmetric matrix since:
\begin{align*}
    \Cmat^T &= \pareT{\Smat+(\mvec-\wvec)(\mvec-\wvec)^T}&& \\
    &= \Smat^T + \pareT{(\mvec-\wvec)(\mvec-\wvec)^T}&& \text{using } \sumTranspose\\
    &= \Smat + (\mvec-\wvec)(\mvec-\wvec)^T&& \text{using } \symetricTranspose,\prodTranspose \\
    &= \Cmat
\end{align*}
The first order differential of the \KLD \wrt $\Vmat$ is given by:
\begin{align*}
      \dd_\Vmat \KLD &= \frac{1}{2}\vvecBT{\Vmatinv-\Vmatinv\pare{\Smat+\avec\avec^T}\Vmatinv}\dd\vvec{\Vmat}&& \\
      &= \frac{1}{2}\pare{\vvecBT{\Vmatinv}\dd\vvec{\Vmat}-\vvecBT{\Vmatinv\Cmat\Vmatinv}\dd\vvec{\Vmat}}&& \text{using \eqn } \ref{vecsum}\\
      &=\frac{1}{2}\pare{\trB{\Vmatinv\dd\Vmat}-\trB{\Vmatinv\Cmat\Vmatinv\dd\Vmat}}&& \text{using \eqns } \ref{dvec},\ref{tr_matprod_2vec},\prodTranspose,\\
      & &&\transposeInverse,\symetricTranspose\\
      &=\frac{1}{2}\trB{\bra{\Vmatinv-\Vmatinv\Cmat\Vmatinv}\dd\Vmat}&& \text{using } \linearTR
\end{align*}
The second differential of the \KLD is obtained by computing the differential of the first differential. Thus:
\begin{align*}
    \dd^2_\Vmat \KLD &= \ddB{\frac{1}{2}\trB{\bra{\Vmatinv-\Vmatinv\Cmat\Vmatinv}\dd\Vmat}}&&\\
    &= \frac{1}{2}\trB{\ddB{\Vmatinv-\Vmatinv\Cmat\Vmatinv}\dd\Vmat + \pare{\Vmatinv-\Vmatinv\Cmat\Vmatinv}\cancelto{0}{\ddB{\dd\Vmat}}}&& \text{using \eqns } \ref{dTR},\ref{d2X},\ref{dprod}\\
    &= \frac{1}{2}\trB{\pare{\ddB{\Vmatinv}-\ddB{\Vmatinv\Cmat\Vmatinv}}\dd\Vmat}&& \text{using \eqn } \ref{dsum}\\
    &= \frac{1}{2}\trB{\pare{-\Vmatinv\dd\Vmat\Vmatinv-\ddB{\Vmatinv}\Cmat\Vmatinv-\Vmatinv\Cmat\ddB{\Vmatinv}}\dd\Vmat}&& \text{using \eqns } \ref{dinv},\ref{dprod},\ref{dAX}\\
    &= \frac{1}{2}\trB{\pare{-\Vmatinv\dd\Vmat\Vmatinv+\Vmatinv\dd\Vmat\Vmatinv\Cmat\Vmatinv+\Vmatinv\Cmat\Vmatinv\dd\Vmat\Vmatinv}\dd\Vmat}&& \text{using \eqn } \ref{dinv}\\
    &= \frac{1}{2}\trB{-\Vmatinv\dd\Vmat\Vmatinv\dd\Vmat+\Vmatinv\dd\Vmat\Vmatinv\Cmat\Vmatinv\dd\Vmat+\Vmatinv\Cmat\Vmatinv\dd\Vmat\Vmatinv\dd\Vmat}&& \\
    &= \frac{1}{2}\trB{-\Vmatinv\dd\Vmat\Vmatinv\dd\Vmat+2\Vmatinv\dd\Vmat\Vmatinv\Cmat\Vmatinv\dd\Vmat}&& \text{using }\linearTR,\\
    & && \circularTR \\
    &= \trB{-\frac{1}{2}\Vmatinv\dd\Vmat\pare{\Vmatinv-2\Vmatinv\Cmat\Vmatinv}\dd\Vmat} \\
    &= \trB{-\frac{1}{2}\Vmatinv\braT{\dd\Vmat}\pare{\Vmatinv-2\Vmatinv\Cmat\Vmatinv}\dd\Vmat}  && \text{using \eqn } \ref{dXT}, \symetricTranspose
\end{align*}
From where the Hessian can be identified:
\begin{align*}
    \frac{\dd\KLD}{\dd \vvecT{\Vmat},\dd\vvec\Vmat}&= \frac{1}{2}\pare{\braT{-\frac{1}{2}\Vmatinv}\otimes\bra{\Vmatinv-2\Vmatinv\Cmat\Vmatinv}+\bra{-\frac{1}{2}\Vmatinv}\otimes\braT{\Vmatinv-2\Vmatinv\Cmat\Vmatinv}}&& \text{using \eqn } \ref{tr_dXdX_2Hess}\\
    &= \frac{1}{2}\pare{-\frac{1}{2}\Vmatinv\otimes\bra{\Vmatinv-2\Vmatinv\Cmat\Vmatinv}-\frac{1}{2}\Vmatinv\otimes\bra{\Vmatinv-2\Vmatinv\Cmat\Vmatinv}}\, \text{using} \\
    & \,\,\,\,\,\,\,\,\,\,\,\,\,\,\,\,\,\,\,\,\,\,\,\,\,\,\,\,\,\,\,\,\,\,\,\,\,\,\,\,\,\,\,\,\,\,\,\,\,\,\,\,\,\,\,\,\,\,\,\,\,\,\,\,\,\,\,\,\,\,\,\,\,\,\,\,\,\,\,\,\,\,\,\,\,\,\,\,\,\,\,\,\,\,\,\,\,\,\,\,\,\,\,\,\,\,\,\,\,\,\,\,\,\,\,\,\,\,\,\,\,\,\,\,\,\,\,\,\,\,\,\,\,\,\,\,\,\,\,\,\,\,\,\,\,\,\,\,\,\,\,\,\,\,\,\,\,\,\,\,\,\, \transposeInverse,\symetricTranspose,\\
    & \,\,\,\,\,\,\,\,\,\,\,\,\,\,\,\,\,\,\,\,\,\,\,\,\,\,\,\,\,\,\,\,\,\,\,\,\,\,\,\,\,\,\,\,\,\,\,\,\,\,\,\,\,\,\,\,\,\,\,\,\,\,\,\,\,\,\,\,\,\,\,\,\,\,\,\,\,\,\,\,\,\,\,\,\,\,\,\,\,\,\,\,\,\,\,\,\,\,\,\,\,\,\,\,\,\,\,\,\,\,\,\,\,\,\,\,\,\,\,\,\,\,\,\,\,\,\,\,\,\,\,\,\,\,\,\,\,\,\,\,\,\,\,\,\,\,\,\,\,\,\,\,\,\,\,\,\,\,\,\,\,\,\sumTranspose,\\
    &\,\,\,\,\,\,\,\,\,\,\,\,\,\,\,\,\,\,\,\,\,\,\,\,\,\,\,\,\,\,\,\,\,\,\,\,\,\,\,\,\,\,\,\,\,\,\,\,\,\,\,\,\,\,\,\,\,\,\,\,\,\,\,\,\,\,\,\,\,\,\,\,\,\,\,\,\,\,\,\,\,\,\,\,\,\,\,\,\,\,\,\,\,\,\,\,\,\,\,\,\,\,\,\,\,\,\,\,\,\,\,\,\,\,\,\,\,\,\,\,\,\,\,\,\,\,\,\,\,\,\,\,\,\,\,\,\,\,\,\,\,\,\,\,\,\,\,\,\,\,\,\,\,\,\,\,\,\,\,\,\,\,\prodTranspose,\\
    &\,\,\,\,\,\,\,\,\,\,\,\,\,\,\,\,\,\,\,\,\,\,\,\,\,\,\,\,\,\,\,\,\,\,\,\,\,\,\,\,\,\,\,\,\,\,\,\,\,\,\,\,\,\,\,\,\,\,\,\,\,\,\,\,\,\,\,\,\,\,\,\,\,\,\,\,\,\,\,\,\,\,\,\,\,\,\,\,\,\,\,\,\,\,\,\,\,\,\,\,\,\,\,\,\,\,\,\,\,\,\,\,\,\,\,\,\,\,\,\,\,\,\,\,\,\,\,\,\,\,\,\,\,\,\,\,\,\,\,\,\,\,\,\,\,\,\,\,\,\,\,\,\,\,\,\,\,\,\,\,\,\,\kronScalar\\
    &= -\frac{1}{2}\Vmatinv\otimes\bra{\Vmatinv-2\Vmatinv\pare{\Smat+(\mvec-\wvec)(\mvec-\wvec)^T}\Vmatinv}&& 
\end{align*}
\cbox{
\textit{\textbf{Differential}}
\begin{align*}
    \dd^2_{\vvec \Vmat} = \braT{\dd\vvec\Vmat}-\frac{1}{2}\Vmatinv\otimes\bra{\Vmatinv-2\Vmatinv\pare{\Smat+(\mvec-\wvec)(\mvec-\wvec)^T}\Vmatinv}\dd\vvec\Vmat
\end{align*}
}
\cbox{
\textit{\textbf{Hessian}}
\begin{align*}
    \frac{\dd\KLD}{\dd \vvecT{\Vmat},\dd\vvec\Vmat} &= -\frac{1}{2}\Vmatinv\otimes\bra{\Vmatinv-2\Vmatinv\pare{\Smat+(\mvec-\wvec)(\mvec-\wvec)^T}\Vmatinv} \in \realNsquareNsquare
\end{align*}
}
where the resulting matrix is the Hessian matrix corresponding to the function with only $\Vmat$ being a variable, and thus is symmetric (as we will check later). Considering only the unique elements from $\Vmat$, we have:
\begin{align*}
     \dd^2_{\Vmat} &= \braT{\dd\vvec\Vmat}-\frac{1}{2}\Vmatinv\otimes\bra{\Vmatinv-2\Vmatinv\pare{\Smat+(\mvec-\wvec)(\mvec-\wvec)^T}\Vmatinv}\dd\vvec\Vmat \\
     &= \braT{\Dn \dd\vvech\Vmat}-\frac{1}{2}\Vmatinv\otimes\bra{\Vmatinv-2\Vmatinv\pare{\Smat+(\mvec-\wvec)(\mvec-\wvec)^T}\Vmatinv}\Dn\dd\vvech\Vmat && \text{using \eqns } \ref{vec2vech},\ref{dAX}\\
     &= \braT{ \dd\vvech\Vmat}\DnT-\frac{1}{2}\Vmatinv\otimes\bra{\Vmatinv-2\Vmatinv\pare{\Smat+(\mvec-\wvec)(\mvec-\wvec)^T}\Vmatinv}\Dn\dd\vvech\Vmat && \text{using } \prodTranspose
\end{align*}
\cbox{
\textit{\textbf{Differential}}
\begin{align*}
\dd^2_{\vvech \Vmat} = \braT{\dd\vvech\Vmat}\DnT-\frac{1}{2}\Vmatinv\otimes\bra{\Vmatinv-2\Vmatinv\pare{\Smat+(\mvec-\wvec)(\mvec-\wvec)^T}\Vmatinv}\Dn\dd\vvech\Vmat
\end{align*}
}
with:
\cbox{
\textit{\textbf{Hessian}}
\begin{align*}
    \frac{\dd\KLD}{\dd \vvechT{\Vmat},\dd\vvech\Vmat} &= \DnT-\frac{1}{2}\Vmatinv\otimes\bra{\Vmatinv-2\Vmatinv\pare{\Smat+(\mvec-\wvec)(\mvec-\wvec)^T}\Vmatinv} \Dn \in \realNDupNDup
\end{align*}
}
\subsubsection{From the second order differential on separate terms and through canonical forms.}
The first order differential of the \KLD \wrt $\Vmat$ is given by:
\begin{align*}
      \dd_\Vmat \KLD &=\frac{1}{2}\trB{\bra{\Vmatinv-\Vmatinv\Cmat\Vmatinv}\dd\Vmat}; \Cmat = \pare{\Smat+\avec\avec^T}  && \\
      &=\frac{1}{2}\pare{\trB{\Vmatinv\dd\Vmat}-\trB{\Vmatinv\Smat\Vmatinv\dd\Vmat}-\trB{\Vmatinv\avec\avec^T\Vmatinv\dd\Vmat}}&& \text{using } \linearTR
\end{align*}
Due to the linearity of the differential, we can now operate on the terms separately towards computing second differentials:
\begin{align*}
\dd_\Vmat\trB{\Vmatinv\dd\Vmat}  &= \\
&= \trB{\ddB{\Vmatinv}\dd\Vmat + \cancelto{0}{\Vmatinv\ddB{\dd\Vmat}}}&& \text{using \eqns } \ref{dprod},\ref{d2X},\ref{dTR}\\
&= \trB{-\Vmatinv\dd\Vmat\Vmatinv\dd\Vmat}&& \text{using \eqn } \ref{dinv} \\
&= \trB{-\Vmatinv\braT{\dd\Vmat}\Vmatinv\dd\Vmat}&& \text{using \eqn } \ref{dXT}, \symetricTranspose
\end{align*}
whereby invoking theorem we have:
\begin{align*}
M^1_\phi  &=   \frac{1}{2}\pare{\braT{-\Vmatinv}\otimes\Vmatinv -\Vmatinv\otimes\braT{\Vmatinv}} && \text{using \eqn } \ref{tr_dXdX_2Hess}\\
&= -\Vmatinv\otimes\Vmatinv&& \text{using } \transposeInverse,\symetricTranspose
\end{align*}
I have used a different notation $M$ since this is not the actual Hessian. On the other side, rather than applying the theorem, we could continue operating to yield the canonical form.
\begin{align*}
\dd_\Vmat\trB{\Vmatinv\dd\Vmat}  &= \hdots \\
&= \trB{-\Vmatinv\dd\Vmat\Vmatinv\dd\Vmat}&& \\
&=-\vvecT{\dd\Vmat}\vvecB{\Vmatinv\dd\Vmat\Vmatinv}&& \text{using \eqns } \ref{tr_matprod_2vec},\ref{dXT},\circularTR,\symetricTranspose\\
&=\braT{\dd{\vvec{\Vmat}}}-\bra{\Vmatinv\otimes\Vmatinv}\dd\vvec{\Vmat}&& \text{using \eqns } \ref{dvec},\ref{vecmat2kron},\transposeInverse,\symetricTranspose
\end{align*}
which coincides with $M^1_\phi$. 

Now operating on the last two terms is equivalent to just changing $\Smat$ by $\avec\avec^T$ in the result. The second differential in this term is:
\begin{align*}
    \dd_\Vmat\pare{\trB{\Vmatinv\Smat\Vmatinv\dd\Vmat}} &= \\
    &= \trB{\ddB{\Vmatinv\Smat\Vmatinv}\dd\Vmat + \Vmatinv\Smat\Vmatinv\cancelto{0}{\ddB{\dd\Vmat}}}&& \text{using \eqns } \ref{dprod},\ref{d2X},\ref{dTR}\\
    &= \trB{\ddB{\Vmatinv}\Smat\Vmatinv\dd\Vmat+\Vmatinv\Smat\ddB{\Vmatinv}\dd\Vmat}&& \text{using \eqns } \ref{dprod},\ref{dAX}\\
    &= \trB{-\Vmatinv\dd\Vmat\Vmatinv\Smat\Vmatinv\dd\Vmat-\Vmatinv\Smat\Vmatinv\dd\Vmat\Vmatinv\dd\Vmat}&& \text{using \eqn } \ref{dinv}\\
    &=\trB{-2\Vmatinv\dd\Vmat\Vmatinv\Smat\Vmatinv\dd\Vmat}&& \text{using } \linearTR,\\
    & &&\circularTR \\
    &=\trB{-2\Vmatinv\braT{\dd\Vmat}\Vmatinv\Smat\Vmatinv\dd\Vmat}&& \text{using \eqn } \ref{dXT},\symetricTranspose
\end{align*}
Invoking theorem yields:
\begin{align*}
    M^2_\phi &= \frac{1}{2}\pare{\braT{-2\Vmatinv}\otimes\bra{\Vmatinv\Smat\Vmatinv}  -2\Vmatinv\otimes\braT{\Vmatinv\Smat\Vmatinv}}&& \text{using \eqn } \ref{tr_dXdX_2Hess}\\
    &= \frac{1}{2}\pare{-2\Vmatinv\otimes\bra{\Vmatinv\Smat\Vmatinv}  -2\Vmatinv\otimes\bra{\Vmatinv\Smat\Vmatinv}}&& \text{using } \transposeInverse,\symetricTranspose\\
    & && \prodTranspose, \kronScalar \\
    &= -2\Vmatinv\otimes\bra{\Vmatinv\Smat\Vmatinv}&& 
\end{align*}
We could however continue operating, to yield the canonical form of the second-order differential:
\begin{align*}
    \dd_\Vmat\pare{\trB{\Vmatinv\Smat\Vmatinv\dd\Vmat}} &= \hdots\\
    &=\trB{-2\Vmatinv\dd\Vmat\Vmatinv\Smat\Vmatinv\dd\Vmat}&& \\
    &=-2\trB{\dd\Vmat\Vmatinv\Smat\Vmatinv\dd\Vmat\Vmatinv}&& \text{using }\circularTR\\
    &=-2\vvecBT{\dd\Vmat}\vvecB{\Vmatinv\Smat\Vmatinv\dd\Vmat\Vmatinv}&& \text{using \eqns } \ref{tr_matprod_2vec},\ref{dXT},\symetricTranspose\\
    &=-2\vvecBT{\dd\Vmat}\Vmatinvt\otimes\bra{\Vmatinv\Smat\Vmatinv}\vvecB{\dd\Vmat}&& \text{using \eqn } \ref{vecmat2kron}\\
    &=\braT{\dd\vvec{\Vmat}}-2\Vmatinv\otimes\bra{\Vmatinv\Smat\Vmatinv}\dd\vvec\Vmat&& \text{using \eqn } \ref{dvec},\transposeInverse,\symetricTranspose
\end{align*}
By analogy, we have:
\begin{align*}
    \dd_\Vmat\pare{\trB{\Vmatinv\avec\avec^T\Vmatinv\dd\Vmat}} &= \hdots \\
    &=\braT{\dd\vvec{\Vmat}}-2\Vmatinv\otimes\bra{\Vmatinv\avec\avec^T\Vmatinv}\dd\vvec\Vmat&& 
\end{align*}
So the total second-order differential is given by:
\begin{align*}
      \dd^2_\Vmat \KLD &=\dd_\Vmat\pare{\frac{1}{2}\trB{\bra{\Vmatinv-\Vmatinv\Cmat\Vmatinv}\dd\Vmat}}; \Cmat = \pare{\Smat+\avec\avec^T}  &&\\
      &=\dd_\Vmat\pare{\frac{1}{2}\pare{\trB{\Vmatinv\dd\Vmat}-\trB{\Vmatinv\Smat\Vmatinv\dd\Vmat}-\trB{\Vmatinv\avec\avec^T\Vmatinv\dd\Vmat}}}&& \text{using }\\
      & && \linearTR\\
      &=\frac{1}{2}\Big(
      \braT{\dd{\vvec{\Vmat}}}-\bra{\Vmatinv\otimes\Vmatinv}\dd\vvec{\Vmat}&& \\
      &-\braT{\dd\vvec{\Vmat}}-2\Vmatinv\otimes\bra{\Vmatinv\Smat\Vmatinv}\dd\vvec\Vmat&&  \\
      &-\braT{\dd\vvec{\Vmat}}-2\Vmatinv\otimes\bra{\Vmatinv\avec\avec^T\Vmatinv}\dd\vvec\Vmat\Big)&& \\
      &= \frac{1}{2}\Big(\braT{\dd{\vvec{\Vmat}}}\Big[-\bra{\Vmatinv\otimes\Vmatinv}\\
      & +2\Vmatinv\otimes\bra{\Vmatinv\Smat\Vmatinv}+2\Vmatinv\otimes\bra{\Vmatinv\avec\avec^T\Vmatinv}\Big] \dd\vvec\Vmat\Big)&& \\
      &= \frac{1}{2}\pare{\braT{\dd{\vvec{\Vmat}}}\bra{-\bra{\Vmatinv\otimes\Vmatinv}+2\Vmatinv\otimes\bra{\Vmatinv\Smat\Vmatinv+\Vmatinv\avec\avec^T\Vmatinv}} \dd\vvec\Vmat}&& \text{using } \\
      & && \kronDistribut\\
      &= \braT{\dd{\vvec{\Vmat}}}\bra{-\frac{1}{2}\bra{\Vmatinv\otimes\Vmatinv}+\Vmatinv\otimes\bra{\Vmatinv\pare{\Smat+\avec\avec^T}\Vmatinv} }\dd\vvec\Vmat&& \\
      &= \braT{\dd{\vvec{\Vmat}}}\bra{-\frac{1}{2}\Vmatinv\otimes\bra{\Vmatinv-2\Vmatinv\pare{\Smat+\avec\avec^T}\Vmatinv}}\dd\vvec\Vmat&& \text{using } \\
      & && \kronDistribut\\
      & && \kronScalar
\end{align*}
From where the Hessian is identified:
\begin{align*}
    \frac{\dd\KLD}{\dd \vvecT{\Vmat},\dd\vvec\Vmat}  = -\frac{1}{2}\Vmatinv\otimes\bra{\Vmatinv-2\Vmatinv\pare{\Smat+(\mvec-\wvec)(\mvec-\wvec)^T}\Vmatinv}
\end{align*}
as expected.
\subsection{Symmetry of the Hessian}
We shall check whether the resulting Hessian matrix is symmetric. We have:
\begin{align*}
    \Hessian{\KLD} = \begin{bmatrix}
\frac{\dd \KLD}{\dd \mvec^2} & \frac{\dd \KLD}{\dd \mvec^T\,\dd\wvec} & \frac{\dd \KLD}{\dd \mvec^T\,\dd\vvec{\Smat}} & \frac{\dd \KLD}{\dd \mvec^T\,\dd\vvec{\Vmat}} \\\\
\frac{\dd \KLD}{\dd \wvec^T\,\dd \mvec} & \frac{\dd \KLD}{\dd \wvec^2} & \frac{\dd \KLD}{\dd \wvec^T\,\dd\vvec{\Smat}} & \frac{\dd \KLD}{\dd \wvec^T\,\dd\vvec{\Vmat}} \\\\
\frac{\dd \KLD}{\dd\vvecT{\Smat}\,\dd \mvec} & \frac{\dd \KLD}{\dd\vvecT{\Smat}\,\dd\wvec} & \frac{\dd \KLD}{\dd\vvec{\Smat}^2} & \frac{\dd \KLD}{\dd\vvecT{\Smat}\,\dd\vvec{\Vmat}} \\\\
\frac{\dd \KLD}{\dd\vvecT{\Vmat}\,\dd \mvec} & \frac{\dd \KLD}{\dd\vvecT{\Vmat}\,\dd\wvec} & \frac{\dd \KLD}{\dd\vvecT{\Vmat}\,\dd\vvec{\Smat}} & \frac{\dd \KLD}{\dd\vvec{\Vmat}^2}
\end{bmatrix} \in \real^{(2N + 2N^2) \times (2N + 2N^2) }
\end{align*}
\begin{align*}
\frac{\dd \KLD}{\dd \mvec^T\,\dd\mvec} &= \Vmatinv \in \realNN \\
\frac{\dd \KLD}{\dd \mvec^T\,\dd\wvec} &= -\Vmatinv  \in \realNN \\
\frac{\dd \KLD}{\dd\mvec^T\,\dd\vvec{\Smat}} &= \zeromat \in \realNNsquare \\
\frac{\dd \KLD}{\dd\mvec^T \dd \vvec\Vmat} &= -\bra{(\mvec-\wvec)^T\Vmatinv}\otimes \Vmatinv \in \realNNsquare \\
\frac{\dd \KLD}{\dd \wvec^T\,\dd\mvec} &= -\Vmatinv \in \realNN \\
\frac{\dd \KLD}{\dd \wvec^T\,\dd\wvec} &= \Vmatinv \in \realNN \\
\frac{\dd \KLD}{\dd\wvec^T\,\dd\vvec{\Smat}} &= \zeromat \in \realNNsquare \\
\frac{\dd \KLD}{\dd\wvec^T \dd \vvec\Vmat} &= \bra{(\mvec-\wvec)^T\Vmatinv} \otimes \Vmatinv \in \realNNsquare \\
\frac{\dd \KLD}{\dd \vvecT{\Smat}\,\dd\mvec} &= \zeromat \in \realNsquareN \\
\frac{\dd \KLD}{\dd \vvecT{\Smat}\,\dd\wvec} &= \zeromat \in \realNsquareN \\
\frac{\dd \KLD}{\dd \vvecT{\Smat}\,\dd\vvec\Smat} &= \frac{1}{2}\bra{\Smatinv\otimes \Smatinv} \in \realNsquareNsquare \\
\frac{\dd \KLD}{\dd\vvecT{\Smat},\dd\vvec{\Vmat}} &= -\frac{1}{2}\pare{\Vmatinv\otimes \Vmatinv}  \in \realNsquareNsquare \\
 \frac{\dd \KLD}{\dd \vvecT{\Vmat}\dd\mvec} &= -\bra{\Vmatinv(\mvec-\wvec)}\otimes\Vmatinv \in \realNsquareN \\
\frac{\dd \KLD}{\dd \vvecT\Vmat\dd\wvec} &= \bra{\Vmatinv(\mvec-\wvec)}\otimes\Vmatinv \in \realNsquareN \\
\frac{\dd \KLD}{\dd\vvecT{\Vmat},\dd\vvec{\Smat}} &= -\frac{1}{2}\pare{\Vmatinv\otimes \Vmatinv } \in \realNsquareNsquare \\
\frac{\dd\KLD}{\dd \vvecT{\Vmat},\dd\vvec\Vmat} &= -\frac{1}{2}\Vmatinv\otimes\bra{\Vmatinv-2\Vmatinv\pare{\Smat+(\mvec-\wvec)(\mvec-\wvec)^T}\Vmatinv} \in \realNsquareNsquare
\end{align*}
We now need to check whether $\Hessian{\KLD} = \Hessian{\KLD}^T$. Since the Hessian is a block matrix, we know that $\blockTranspose$, we perform the following checks:
\begin{flalign*}
       \braT{\frac{\dd \KLD}{\dd \mvec^T\,\dd\mvec}} &= \braT{\Vmatinv} && \\
       &= \Vmatinv && \text{using } \symetricTranspose,\transposeInverse \\
       &= \frac{\dd \KLD}{\dd \mvec^T\,\dd\mvec}
\end{flalign*}
\begin{flalign*}
    \braT{\frac{\dd \KLD}{\dd \wvec^T\,\dd\mvec}} &= \braT{-\Vmatinv}  \\
    &= -\Vmatinv && \text{using } \symetricTranspose,\transposeInverse \\
    &= \frac{\dd \KLD}{\dd \mvec^T\,\dd\wvec}
\end{flalign*}
\begin{flalign*}
    \braT{\frac{\dd \KLD}{\dd \vvecT{\Smat}\,\dd\mvec}} &= \braT{\zeromat} && \\
    &= \frac{\dd \KLD}{\dd\mvec^T,\dd \vvec{\Smat}} &&
\end{flalign*}
\begin{flalign*}
    \braT{\frac{\dd \KLD}{\dd \vvecT{\Vmat}\dd\mvec}} &= \braT{-\bra{\Vmatinv(\mvec-\wvec)}\otimes\Vmatinv}\\
    &\braT{-\bra{\Vmatinv(\mvec-\wvec)}}\otimes\braT{\Vmatinv} && \text{using } \kronTranspose\\
    &-\bra{\pareT{\mvec-\wvec}\Vmatinv}\otimes \Vmatinv && \text{using } \prodTranspose,\symetricTranspose,\transposeInverse\\
    &= \frac{\dd \KLD}{\dd\mvec^T\dd \vvec{\Vmat}}
\end{flalign*}
\begin{flalign*}
    \braT{\frac{\dd \KLD}{\dd \wvec^T\,\dd\wvec}} &= \braT{\Vmatinv} && \\
    &= \Vmatinv && \text{using } \symetricTranspose,\transposeInverse \\
       &= \frac{\dd \KLD}{\dd \wvec^T\,\dd\wvec}
\end{flalign*}
\begin{flalign*}
    \braT{\frac{\dd \KLD}{\dd \vvecT{\Smat}\,\dd\wvec}} &= \braT{\zeromat} && \\
    &= \frac{\dd \KLD}{\dd\wvec^T,\dd \vvec{\Smat}} &&
\end{flalign*}
\begin{flalign*}
     \braT{\frac{\dd \KLD}{\dd \vvecT{\Vmat}\dd\wvec}} &= \braT{\bra{\Vmatinv(\mvec-\wvec)}\otimes\Vmatinv} && \\
    &\braT{\bra{\Vmatinv(\mvec-\wvec)}}\otimes\braT{\Vmatinv} && \text{using } \kronTranspose\\
    &\bra{\pareT{\mvec-\wvec}\Vmatinv}\otimes \Vmatinv && \text{using } \prodTranspose,\symetricTranspose,\transposeInverse\\
    &= \frac{\dd \KLD}{\dd\wvec^T\dd \vvec{\Vmat}}
\end{flalign*}
\begin{flalign*}
    \braT{\frac{\dd \KLD}{\dd \vvecT{\Smat}\,\dd\vvec\Smat}} &= \braT{\frac{1}{2}\bra{\Smatinv\otimes \Smatinv}} && \\
    &= \frac{1}{2}\bra{\braT{\Smatinv}\otimes \braT{\Smatinv}} && \text{using } \kronTranspose\\
    &= \frac{1}{2}\bra{\Smatinv\otimes \Smatinv} && \text{using } \symetricTranspose,\transposeInverse\\
    &= \frac{\dd \KLD}{\dd \vvecT{\Smat}\,\dd\vvec\Smat}
\end{flalign*}
\begin{flalign*}
    \braT{\frac{\dd \KLD}{\dd\vvecT{\Vmat},\dd\vvec{\Smat}}} &= \braT{-\frac{1}{2}\bra{\Vmatinv\otimes \Vmatinv}}  && \\
    &= -\frac{1}{2}\bra{\braT{\Vmatinv}\otimes \braT{\Vmatinv}} && \text{using } \kronTranspose\\
    &= -\frac{1}{2}\bra{\Vmatinv\otimes \Vmatinv} && \text{using } \symetricTranspose,\transposeInverse\\
    &= \frac{\dd \KLD}{\dd \vvecT{\Smat}\,\dd\vvec\Vmat}
\end{flalign*}
\begin{flalign*}
    \braT{\frac{\dd\KLD}{\dd \vvecT{\Vmat},\dd\vvec\Vmat}} &= \braT{-\frac{1}{2}\Vmatinv\otimes\bra{\Vmatinv-2\Vmatinv\pare{\Smat+(\mvec-\wvec)(\mvec-\wvec)^T}\Vmatinv}} && \\
    &= -\frac{1}{2}\braT{\Vmatinv}\otimes\braT{\Vmatinv-2\Vmatinv\pare{\Smat+(\mvec-\wvec)(\mvec-\wvec)^T}\Vmatinv}&& \text{using } \kronTranspose\\
    &= -\frac{1}{2}\braT{\Vmatinv}\otimes\bra{\braT{\Vmatinv}-2\braT{\Vmatinv\pare{\Smat+(\mvec-\wvec)(\mvec-\wvec)^T}\Vmatinv}}&& \text{using } \sumTranspose\\
    &= -\frac{1}{2}\Vmatinv\otimes\bra{\Vmatinv-2\Vmatinv\pare{\Smat+(\mvec-\wvec)(\mvec-\wvec)^T}\Vmatinv}&& \text{using } \prodTranspose,\\
    & &&\sumTranspose\\
    & &&\symetricTranspose,\transposeInverse\\
    &= \frac{\dd\KLD}{\dd \vvecT{\Vmat},\dd\vvec\Vmat}
\end{flalign*}
A similar procedure shows that the Hessian on the unique elements, \ie when using $\vvech{}$, is also a symmetric matrix by using $\prodTranspose$, which moves $\Dn$ and $\DnT$ from left to right and vice versa.
\subsection{A note on performing derivations that yield symmetrical Hessian matrices}

When computing some of the partial derivatives in the Hessian, I found that sometimes it is necessary to work out derivations in a particular way, such that one can finally demonstrate symmetry in the Hessian.

In particular, we have previously shown that:
\begin{align*}
    \frac{\dd \KLD}{\dd\mvec^T \dd \vvec\Vmat} = -\bra{(\mvec-\wvec)^T\Vmatinv}\otimes \Vmatinv \in \realNNsquare
\end{align*}
However, in the first attempt, I obtained:
\begin{align*}
    \dd_\mvec\dd_\Vmat &= \dd_\mvec\bra{\frac{1}{2}\bra{\cancelto{0}{\trB{\Vmatinv\dd\Vmat}}  - \cancelto{0}{\trB{\Vmatinv\Smat\Vmatinv\dd\Vmat}} - \trB{\Vmatinv\avec\avec^T\Vmatinv\dd\Vmat} }}; \avec=(\mvec-\wvec) && \text{using \eqns} \ref{dconst},\ref{dsum}\\
    &= -\frac{1}{2}\bra{\trB{\dd\bra{\avec^T\Cmat\avec}}}; \Cmat = \Vmatinv\dd\Vmat\Vmatinv && \text{using \eqn }  \ref{dTR}, \circularTR\\
    &= -\trB{\avec^T\Cmat\dd\avec} && \text{using \eqns } \ref{dprod},\ref{dXT},\ref{dAX}, \quadraticScalar,\\
    & && \scalarTranspose,\\
    & && \prodTranspose \\
    & && \symetricTranspose, \transposeInverse \\
    &= -\trB{\dd\mvec(\mvec-\wvec)^T\Vmatinv\dd\Vmat\Vmatinv}  && \text{using \eqns }\ref{dsum},\ref{dconst},\\
    & && \circularTR \\
    &= -\braT{\dd \mvec}\vvecB{(\mvec-\wvec)^T\Vmatinv\dd\Vmat\Vmatinv} &&  \text{using \eqns } \ref{tr_matprod_2vec},\ref{vec_of_a_vec} \\
    &= \braT{\dd \mvec}-\Vmatinv \otimes \bra{(\mvec-\wvec)^T\Vmatinv}\dd\vvec\Vmat && \text{using \eqns } \ref{vecmat2kron},\ref{dvec},\\
    & && \transposeInverse, \symetricTranspose
\end{align*}
from where:
\begin{align*}
    \frac{\dd \KLD}{\dd\mvec^T \dd \vvec\Vmat} = -\Vmatinv \otimes \bra{(\mvec-\wvec)^T\Vmatinv} \in \realNNsquare
\end{align*}
However, when checking the symmetry of the Hessian, we found that showing the transpose of this result was equal to:
\begin{align*}
    \frac{\dd \KLD}{\dd \vvecT{\Vmat}\dd\mvec} &= -\bra{\Vmatinv(\mvec-\wvec)}\otimes\Vmatinv \in \realNsquareN 
\end{align*}
was not possible, or at least I did not figure it out. Up to this point, I found that this differential was pretty similar to that in \cite{minka} equation $(123)$, which yields the partial derivative given by equation $(126)$. This allowed me to ensure that my derivation of $\frac{\dd \KLD}{\dd \vvecT{\Vmat}\dd\mvec}$ was in the correct way, and that some work needed to be done with $\frac{\dd \KLD}{\dd\mvec^T \dd \vvec\Vmat}$ to yield an expression whose transposed was equal. This was finally achieved by carefully working on the differential and is the main result shown in \usec \ref{dmdv_derivation}.

Later on, I found in \cite{magnus99} that:
\begin{align*}
    \vvecB{\Amat\Bmat\dvec} = \bra{\dvec^T \otimes \Amat } \vvec \Bmat = \bra{\Amat \otimes \dvec^T} \vvec{\Bmat^T}
\end{align*}
If we apply this result to the derivation in this section, we can show that:
\begin{align*}
    \hdots &= -\braT{\dd \mvec}\vvecB{(\mvec-\wvec)^T\Vmatinv\dd\Vmat\Vmatinv} \\
    &= -\braT{\dd \mvec}\vvecB{\Vmatinv\dd\Vmat\Vmatinv(\mvec-\wvec)} && \text{using \eqns } \ref{vec_of_a_vec},\ref{dXT},\prodTranspose, \\
    & && \transposeInverse, \symetricTranspose\\
    &= \braT{\dd \mvec}-\bra{(\mvec-\wvec)^T\Vmatinv}\otimes \Vmatinv\dd\vvec\Vmat && \text{using }  \vvecB{\Amat\Bmat\dvec} = \bra{\dvec^T \otimes \Amat } \vvec \Bmat\\
    & && \prodTranspose, \transposeInverse, \symetricTranspose, \text{\eqn } \ref{dvec}\\
    &= \braT{\dd \mvec} -\Vmatinv \otimes \bra{(\mvec-\wvec)^T\Vmatinv}\dd\vvec\Vmat && \text{using } \vvecB{\Amat\Bmat\dvec} = \bra{\Amat \otimes \dvec^T} \vvec{\Bmat^T} \\
    & && \prodTranspose, \transposeInverse, \symetricTranspose, \text{\eqns } \ref{dvec},\ref{dXT}
\end{align*}
showing that both expressions were valid derivations of the partial derivative $\frac{\dd \KLD}{\dd\mvec^T \dd \vvec\Vmat}$.

Overall this shows how different procedures can lead to different expressions where some properties, such as symmetry or equivalence, might be difficult to prove; and how knowing different versions of the relationship between different operators such as de trace, vec or Kronecker product might simplify derivations towards finding expressions that satisfy different properties (symmetry in this case).

Note that since $\kronTranspose$ then proving that:
\begin{align*}
  \braT{-\bra{\Vmatinv(\mvec-\wvec)}\otimes\Vmatinv}  =  -\Vmatinv \otimes \bra{(\mvec-\wvec)^T\Vmatinv} 
\end{align*}
is, at least for me, nontrivial; while proving that:
\begin{align*}
   \braT{-\bra{\Vmatinv(\mvec-\wvec)}\otimes\Vmatinv}  = -\bra{(\mvec-\wvec)^T\Vmatinv}\otimes \Vmatinv
\end{align*}
is nearly direct. On the other hand, checking that both partial derivative expressions are equal:
\begin{align*}
    -\bra{(\mvec-\wvec)^T\Vmatinv}\otimes \Vmatinv = -\Vmatinv \otimes \bra{(\mvec-\wvec)^T\Vmatinv}
\end{align*}
is, again with my knowledge, non-trivial.

\clearpage
%\bibliographystyle{apalike}
%\bibliography{main}

\end{document}